\documentclass[a4paper,fleqn]{cas-dc}

\usepackage[numbers,sort&compress]{natbib}
\usepackage{tabularx}
\usepackage{array}
\usepackage{graphicx}
\usepackage{placeins}

\begin{document}
\let\WriteBookmarks\relax

\shorttitle{Retention Dynamics in Fine-Tuned Image Classifiers}

\title[mode = title]{Not All Forgetting Is Equal: Retention Dynamics in Fine-Tuned Image Classifiers}

%

\author[1]{Miit Daga}[
    orcid=0009-0005-4629-458X
]
\ead{miit.daga2022@vitstudent.ac.in}

\credit{Conceptualization, Methodology, Software, Validation, Formal analysis, Investigation, Writing - Original Draft}

\author[1]{Swarna Priya Ramu}[
    orcid=0000-0002-8287-9690
]
\cormark[1]
\ead{swarnapriya.rm@vit.ac.in}

\credit{Supervision, Writing - Original Draft, Writing - Review \& Editing, Project administration}

\affiliation[1]{organization={School of Computer Science Engineering and Information Systems, Vellore Institute of Technology},
    city={Vellore},
    postcode={632014},
    state={Tamil Nadu},
    country={India}}

\cortext[cor1]{Corresponding author: Swarna Priya Ramu}


\begin{abstract}
Fine-tuning a pretrained classifier leaves some samples reliably learned and others cycling between correct and incorrect. Curriculum learning, data pruning and dataset cartography assume that pattern is a property of the sample, untested. We record per-sample correctness at every epoch while fine-tuning ResNet-18 and DeiT-Small on an imbalanced retinal OCT dataset and CUB-200-2011, matching samples by image identity and holding the split fixed across seeds. Per-sample retention is reproducible: cross-run Spearman $\rho$ of the fitted decay constant is $0.37$ to $0.59$ over ten seeds. It is architecture-specific: two runs of one backbone agree more than two backbones on identical data ($0.45$ and $0.59$ within against $0.30$ between on OCTDL). Loss after five frozen-backbone epochs predicts a \textit{different} run's decay constant at $\rho = 0.29$ to $0.43$. The Ebbinghaus exponential does not survive: monotone decay, the one shape it can represent, is $0.1\%$ to $0.8\%$ of samples, and on traces that do forget mean $R^2$ is negative in all four configurations. A power law and a free-asymptote variant fail on the same traces: the defect is monotonicity. Across five sampling arms with matched exposure, prioritisation ratios of $2.7\times$ to $28\times$, and an online variant, none of 48 comparisons against uniform sampling survives Benjamini-Hochberg correction, though three seeds detect only about four accuracy points. A stable, cheap difficulty score does not buy generalisation through sampling. Patient-grouped splitting, the remedy for a leak reaching 76\% to 78\% of OCT test images, moves that dataset's headline metrics by less than their run-to-run spread.
\end{abstract}


\begin{highlights}
\item Per-sample forgetting is reproducible across seeds (rho 0.37 to 0.59)
\item CNNs and ViTs forget different samples on identical data
\item The shape the exponential model needs occurs in 0.1\% to 0.8\% of samples
\item Loss after 5 warmup epochs predicts decay in another run (rho 0.29-0.43)
\item No sampling arm beats uniform across a 28-fold prioritisation sweep
\end{highlights}

\begin{keywords}
forgetting dynamics \sep fine-tuning \sep sample difficulty \sep CNN versus ViT \sep spaced repetition
\end{keywords}

\maketitle
\makeatletter
\renewenvironment{figure}[1][tbp]{%
  \@float{figure}[#1]%
}{%
  \end@float
}
\renewenvironment{figure*}[1][tbp]{%
  \@dblfloat{figure}[#1]%
}{%
  \end@dblfloat
}
\makeatother
\renewcommand{\floatpagefraction}{0.6}
\renewcommand{\textfraction}{0.1}
\renewcommand{\topfraction}{0.9}
\renewcommand{\bottomfraction}{0.8}
\renewcommand{\dbltopfraction}{0.9}
\renewcommand{\dblfloatpagefraction}{0.6}
\setcounter{topnumber}{3}
\setcounter{bottomnumber}{2}
\setcounter{totalnumber}{5}
\setcounter{dbltopnumber}{2}

\section{Introduction}\label{sec:intro}

In 1885 Hermann Ebbinghaus measured how quickly he forgot lists of nonsense syllables and reported retention that dropped sharply at first, then levelled off~\cite{ebbinghaus1885}. Which functional form best describes that curve is still argued over, and replication work fits the data with more than one candidate~\cite{murre2015replication}. What survives the argument is the weaker claim we borrow here: forgetting has a rate, the rate differs between items, and it can be measured.

Deep networks do something that looks similar. During optimisation a training example can be classified correctly at one epoch and incorrectly at the next. Toneva et al.~\cite{toneva2018empirical} documented this for training from scratch on CIFAR and found that many samples, once learned, are never forgotten. Fine-tuning is a different regime. Pretrained representations constrain the loss landscape, so most samples are learned within a few epochs and only a subset keeps cycling.

Whether that subset is the same subset from run to run is not a decorative question. Several methods depend on it. Curriculum learning~\cite{bengio2009curriculum} orders samples from easy to hard, self-paced learning~\cite{kumar2010self} lets the model pick that order, dataset cartography~\cite{swayamdipta2020dataset} sorts samples into easy, ambiguous, and hard regions, and data pruning~\cite{paul2021deep} uses early-training gradient magnitudes to decide what to discard. Each computes a per-sample score from one training trajectory and then acts on it. That only works if the score transfers.

We measure whether it does. We record per-sample correctness at the end of every epoch while fine-tuning ResNet-18~\cite{he2016deep} and DeiT-Small~\cite{touvron2021training} on a heavily imbalanced retinal OCT dataset and on CUB-200-2011, fit a decay constant $\lambda_i$ to each sample's retention trace, and compare those constants across architectures, seeds, and classes. Two design choices do most of the work. Samples are matched by image identity rather than by position in the training array, which matters because a different seed produces a different split. And a factorial condition holds the split fixed so that the seed moves only initialisation, augmentation, and batch order, which separates training stochasticity from data composition. Our contributions:

\begin{enumerate}
    \item \textbf{Per-sample retention is reproducible.} Cross-run Spearman correlation of $\lambda$ is $0.37$ to $0.59$ across two datasets and two architectures, measured with the split held fixed over ten seeds. Difficulty scores do transfer between runs, contrary to what a naive index-based comparison suggests.
    \item \textbf{Retention is architecture-specific.} Two runs of the same architecture agree more than two different architectures on identical data ($0.45$ and $0.59$ within, $0.30$ between, on OCTDL). The same ordering holds on CUB-200.
    \item \textbf{The exponential model does not describe the traces it is fitted to.} Restricted to samples that actually forget at least once, mean $R^2$ is between $-0.08$ and $0.00$. Reported fit quality in this setting is largely a count of samples that never forgot, not a measure of fit. Sorted by shape instead, monotone decay, the one case the model is built for, is $0.1\%$ to $0.8\%$ of samples in every configuration. A free-asymptote exponential and a power law fail on the same traces, so what fails is monotonicity rather than the particular curve.
    \item \textbf{Loss after head warmup predicts retention in a different run.} Spearman $\rho$ between Phase~1 loss in one run and $\lambda$ in another is $0.29$ to $0.43$, so five epochs of warmup flag vulnerable samples without a full training run.
    \item \textbf{Scheduling on $\lambda$ does not improve accuracy.} Across five sampling arms with matched exposure, a temperature sweep spanning realised prioritisation ratios of $2.7\times$ to $28\times$, and an online variant that re-estimates $\lambda$ during training, no comparison against uniform sampling survives Benjamini-Hochberg correction, though three seeds per arm bound what that rules out (Section~\ref{sec:limitations}).
\end{enumerate}

The last two results sit together awkwardly. A stable, cheaply estimated signal, yet acting on it changes nothing, so an unstable target is not the explanation for why sample-level scheduling underdelivers. That leaves the architecture comparison and the early predictor as the measurements this paper adds, and the sampler null as the boundary they place on using them.

We also report a measurement problem specific to medical imaging. Splitting OCTDL at the image level, as is common, places roughly $77\%$ of test images in the same patient as a training image. We split at the patient level throughout and quantify what the naive split costs, which turns out to be less than the scale of the leak would suggest.

Sections~\ref{sec:related} and~\ref{sec:background} cover related work and the technical background. Section~\ref{sec:method} describes retention tracking, decay fitting, and the sampler, and Section~\ref{sec:setup} the datasets, splitting, and statistical protocol. Results and discussion follow in Section~\ref{sec:results}, and Section~\ref{sec:conclusion} closes with limitations and future work. Appendix~\ref{app:perclass} holds per-class test metrics and Appendix~\ref{app:stats} the full statistical battery.

\section{Related Work}\label{sec:related}

Three lines of work bear on this paper: measurements of how individual samples behave during training, methods that schedule training on those measurements, and comparisons of what CNNs and transformers represent.

\subsection{Per-sample learning dynamics}\label{sec:rel_dynamics}

Toneva et al.~\cite{toneva2018empirical} counted forgetting events, transitions from correct to incorrect, while training from scratch on CIFAR. They found a large set of examples that are never forgotten once learned and showed that those examples can be removed without hurting accuracy. Swayamdipta et al.~\cite{swayamdipta2020dataset} tracked confidence and its variance across epochs and used the two to sort a dataset into easy, ambiguous, and hard regions. Paul et al.~\cite{paul2021deep} took a cheaper route, scoring each sample by the norm of its gradient or its error vector after only a few epochs, and pruned on that score. Sorscher et al.~\cite{sorscher2022beyond} pushed the same idea to scale, showing that a good difficulty score lets pruning beat power-law scaling, which raises the stakes on whether such a score is stable. Feldman~\cite{feldman2020does} gives a reason to expect some of it not to be: on long-tailed data, memorising rare examples is necessary for good generalisation, so a high difficulty score may mark an example the model must memorise rather than one it can discard. Pleiss et al.~\cite{pleiss2020identifying} use the area under the margin to separate those cases from mislabelled data. Maini et al.~\cite{maini2022characterizing} used a second training split to separate genuinely rare examples from mislabelled ones, which event counts alone conflate.

All four measure a sample once and act on the number. None reports what happens to that number when the run is repeated, and all four study training from scratch.

\subsection{Scheduling on per-sample scores}\label{sec:rel_sched}

Curriculum learning~\cite{bengio2009curriculum} presents easy samples first and hard ones later; self-paced learning~\cite{kumar2010self} lets the model decide which are which as it goes. Hacohen and Weinshall~\cite{hacohen2020let} report that networks learn examples in a broadly similar order across architectures and initialisations, which would make such scores portable. Their evidence comes from training from scratch, so whether the ordering survives fine-tuning, where pretrained weights already constrain the loss landscape, is a separate question.

A second family borrows directly from human memory research. Settles and Meeder~\cite{settles2016trainable} fit half-life regression to Duolingo review logs and scheduled practice from the fitted half-lives. Amiri et al.~\cite{amiri2017repeat} carried spaced repetition into neural network training on NLP tasks, spacing example presentations by how quickly the model appeared to forget them. Kline~\cite{kline2025human} comes closest to the question we ask. He tracks a class-level recall probability, the softmax-normalised similarity between a multilayer perceptron's hidden state and stored class prototypes, after removing one MNIST class from training altogether, and reports that it decays along a power law and recovers under spaced review. Two things differ from our setting. The unit is the class rather than the sample, and an aggregate over a class can fall smoothly while the binary traces beneath it oscillate. And the forgetting is induced by withholding data rather than observed during ordinary fine-tuning, so his reviews reintroduce examples the model has stopped seeing, where ours only change how often samples that are always present get drawn. None of these three lines has been tested on image fine-tuning, and none reports what the realised sampling distribution looked like once the schedule was applied.

Active learning offers the closest precedent for what we find. Munjal et al.~\cite{munjal2022towards} and Mittal et al.~\cite{mittal2019parting} both report that sophisticated acquisition functions stop beating random selection once the comparison controls for training budget, augmentation, and regularisation. The mechanism they diagnose is not that the acquisition scores are meaningless but that the gains attributed to them came from elsewhere in the pipeline. Our sampler result is the same shape of finding in a different setting, and the two together are consistent with the same failure mode appearing outside acquisition.

\subsection{Architectural differences in representation}\label{sec:rel_arch}

Raghu et al.~\cite{raghu2021vision} compared ViTs and CNNs layer by layer and found transformers keep more uniform representations across depth, with earlier access to global context. That work compares aggregate representational structure, as does the broader literature on how transformers and convolutional networks differ in what they learn~\cite{touvron2021training}. Whether two architectures also disagree about which individual training samples they hold onto is not something layer-wise similarity answers.

\subsection{Research gaps and our position}\label{sec:rel_gap}

Table~\ref{tab:related} places this paper against those lines. Three gaps remain open. Per-sample difficulty is measured but its run-to-run reproducibility is not; the exponential form borrowed from memory research is fitted but not validated against the traces it is applied to; and spaced-repetition schedulers are proposed without reporting whether the resulting sampling distribution differs from uniform. We address all three, and our answer to the first agrees with Hacohen and Weinshall rather than contradicting them: the ordering does largely carry over, in the fine-tuning regime as well.

\begin{table}[!h]
\centering
\caption{Position of this work against prior lines. $\checkmark$ = addressed, $\sim$ = partially, $\times$ = not addressed.}\label{tab:related}
\resizebox{\columnwidth}{!}{
\begin{tabular}{|l|c|c|c|c|c|}
\hline
\textbf{Work} & \textbf{Fine-} & \textbf{Per-sample} & \textbf{Run-to-run} & \textbf{Instance-level} & \textbf{Schedule} \\
 & \textbf{tuning} & \textbf{decay model} & \textbf{stability} & \textbf{arch. compare} & \textbf{tested} \\
\hline
Toneva et al.~\cite{toneva2018empirical} & $\times$ & $\times$ & $\times$ & $\times$ & $\sim$ \\
\hline
Swayamdipta et al.~\cite{swayamdipta2020dataset} & $\times$ & $\times$ & $\times$ & $\times$ & $\times$ \\
\hline
Paul et al.~\cite{paul2021deep} & $\times$ & $\times$ & $\times$ & $\times$ & $\sim$ \\
\hline
Maini et al.~\cite{maini2022characterizing} & $\times$ & $\times$ & $\sim$ & $\times$ & $\times$ \\
\hline
Hacohen \& Weinshall~\cite{hacohen2020let} & $\times$ & $\times$ & $\checkmark$ & $\sim$ & $\times$ \\
\hline
Settles \& Meeder~\cite{settles2016trainable} & $\times$ & $\checkmark$ & $\times$ & $\times$ & $\checkmark$ \\
\hline
Amiri et al.~\cite{amiri2017repeat} & $\times$ & $\checkmark$ & $\times$ & $\times$ & $\checkmark$ \\
\hline
Kline~\cite{kline2025human} & $\times$ & $\times$ & $\times$ & $\times$ & $\checkmark$ \\
\hline
Raghu et al.~\cite{raghu2021vision} & $\sim$ & $\times$ & $\times$ & $\sim$ & $\times$ \\
\hline
Sorscher et al.~\cite{sorscher2022beyond} & $\times$ & $\times$ & $\times$ & $\times$ & $\sim$ \\
\hline
Munjal et al.~\cite{munjal2022towards} & $\sim$ & $\times$ & $\sim$ & $\times$ & $\checkmark$ \\
\hline
Mittal et al.~\cite{mittal2019parting} & $\sim$ & $\times$ & $\sim$ & $\times$ & $\checkmark$ \\
\hline
\textbf{This work} & $\checkmark$ & $\checkmark$ & $\checkmark$ & $\sim$ & $\checkmark$ \\
\hline
\end{tabular}
}
\end{table}

\section{Background and Fundamentals}\label{sec:background}

The measurements in this paper rest on four things: two datasets that stress difficulty in different ways, a fine-tuning protocol split into two phases, a definition of forgetting narrower than the word usually implies, and a small set of statistics for reading rank agreement and set overlap. This section sets out each. Two of them carry a default that reads naturally and gives the wrong answer, and both return in Section~\ref{sec:results}: two OCT images from one patient are not independent observations, and a Jaccard overlap means little until it is set against what two unrelated selections would produce anyway.

\subsection{Retinal OCT and the unit of analysis}\label{sec:bg_oct}

Fine-tuning a pretrained classifier is the default approach in medical imaging, where labelled data are scarce~\cite{kim2022transfer}. Optical coherence tomography produces cross-sectional images of the retina, and a single imaging session yields several B-scans of the same eye. OCTDL~\cite{kulyabin2024octdl} labels each image with one of seven categories: age-related macular degeneration (AMD), diabetic macular edema (DME), epiretinal membrane (ERM), normal retina (NO), retinal artery occlusion (RAO), retinal vein occlusion (RVO), and vitreomacular interface disease (VID). Class sizes are extremely uneven, with roughly a 56:1 ratio between the largest and smallest over the full dataset, and about 45:1 within the training split once patients are partitioned, as the per-class sizes in Table~\ref{tab:octdl_class} show.

The detail that matters for evaluation is that images are not independent. In OCTDL the filename encodes a patient identifier, and 2{,}064 images come from 820 patients, an average of 2.5 images each and up to 10 for one patient. Two B-scans of the same eye share anatomy, acquisition settings, and pathology. Splitting at the image level therefore places near-duplicates of a training image into the test set. This failure mode is well documented in medical imaging, where repeated acquisitions from one subject spread across partitions have produced optimistic results often enough to be catalogued as a standard pitfall~\cite{roberts2021common,degrave2021ai,varoquaux2022machine}. The unit of analysis has to be the patient, not the image. What is less often measured is how much the choice actually changes the reported numbers, which Section~\ref{sec:res_char} does for this dataset.

\subsection{Fine-grained recognition as a contrasting regime}\label{sec:bg_cub}

CUB-200-2011~\cite{wah2011caltech} contains 11{,}788 photographs of 200 bird species, roughly balanced at 41 to 60 images per class. Where OCTDL stresses class imbalance~\cite{johnson2019survey}, CUB stresses inter-class similarity: many species differ only in plumage detail or bill shape. Each image is an independent photograph, so the image is the correct unit and no grouping is required. Using both datasets lets the same measurement run against two different sources of difficulty.

\subsection{Two-phase fine-tuning}\label{sec:bg_finetune}

Fine-tuning a pretrained classifier on a small dataset is usually split into two phases. The backbone is frozen first and only the new classification head is trained, which lets the randomly initialised head reach a sensible operating point without large gradients disturbing the pretrained features. The backbone is then unfrozen and the whole network is trained at a lower learning rate. We follow this protocol with ResNet-18~\cite{he2016deep} and DeiT-Small~\cite{touvron2021training} initialised from ImageNet-1K weights through \texttt{timm}~\cite{rw2019timm}.

The consequence for retention is that the two phases behave differently. During the frozen phase most samples are either learned or not, and little changes. Once the backbone unfreezes, the representation moves, and samples that were classified correctly can stop being classified correctly.

\subsection{Forgetting events and retention traces}\label{sec:bg_forgetting}

Following Toneva et al.~\cite{toneva2018empirical}, a \textit{forgetting event} for sample $i$ occurs at epoch $e$ when the sample was classified correctly at $e-1$ and incorrectly at $e$. Recording correctness at every epoch gives each sample a binary \textit{retention trace}, and stacking the traces gives a matrix over samples and epochs.

One point of terminology matters here, because the word is overloaded. In continual and lifelong learning, forgetting means interference between tasks trained in sequence, where competence on an earlier task degrades as a later one is learned~\cite{mccloskey1989catastrophic,kirkpatrick2017overcoming}. Nothing in this paper is sequential: there is one task, one dataset, and one training run, and forgetting means that a single training image classified correctly at one epoch is classified incorrectly at the next. The two phenomena share a name and little else.

Ebbinghaus-style analysis summarises such a trace by a decay rate~\cite{ebbinghaus1885}. That summary carries an assumption worth stating early: a decay curve is monotone, so it can represent a sample that is learned and then progressively lost, but not one that is forgotten and later relearned. The psychological literature has never settled on a single functional form~\cite{murre2015replication}, and Section~\ref{sec:res_char} reports how well the exponential actually describes the traces here.

\subsection{Metrics and statistical tests}\label{sec:bg_stats}

Classification is reported with top-1 accuracy, balanced accuracy, macro and weighted F1, Cohen's $\kappa$, and Matthews correlation coefficient, with per-class precision, recall, and specificity in Appendix~\ref{app:perclass}. Accuracy alone is misleading on OCTDL because the majority class covers most of the data, so macro-F1 and balanced accuracy carry the per-class picture.

Rank agreement between two runs is measured with Spearman's $\rho$, which is appropriate because the decay constants are heavily tied at zero and their absolute scale is not meaningful. Ties take mid-ranks throughout. Because never-learned samples are held out as their own category rather than given an imputed value (Section~\ref{sec:decay}), the tied mass sits at genuine zeros and not at an artificial ceiling. Set overlap between two top-$k$ selections is measured with the Jaccard index, which has a non-zero expectation under independence: for two independent top-$k$ fractions the expected value is $k/(2-k)$, so $k=10\%$ gives $0.053$ and $k=50\%$ gives $0.333$. Any observed overlap must be read against that baseline.

Comparisons between sampling arms use paired tests on the seed as the unit, and McNemar's test~\cite{mcnemar1947note} on the shared test set when a higher-powered sample-level comparison is available; Section~\ref{sec:setup_stats} gives the full protocol and says which comparisons fall outside it. Because each family involves many comparisons, all reported $p$-values are corrected with the Benjamini-Hochberg procedure~\cite{benjamini1995controlling}, and effect sizes and bootstrap intervals are given alongside them.

\section{Methodology}\label{sec:method}

\begin{figure*}[!h]
  \centering
  \includegraphics[width=\linewidth]{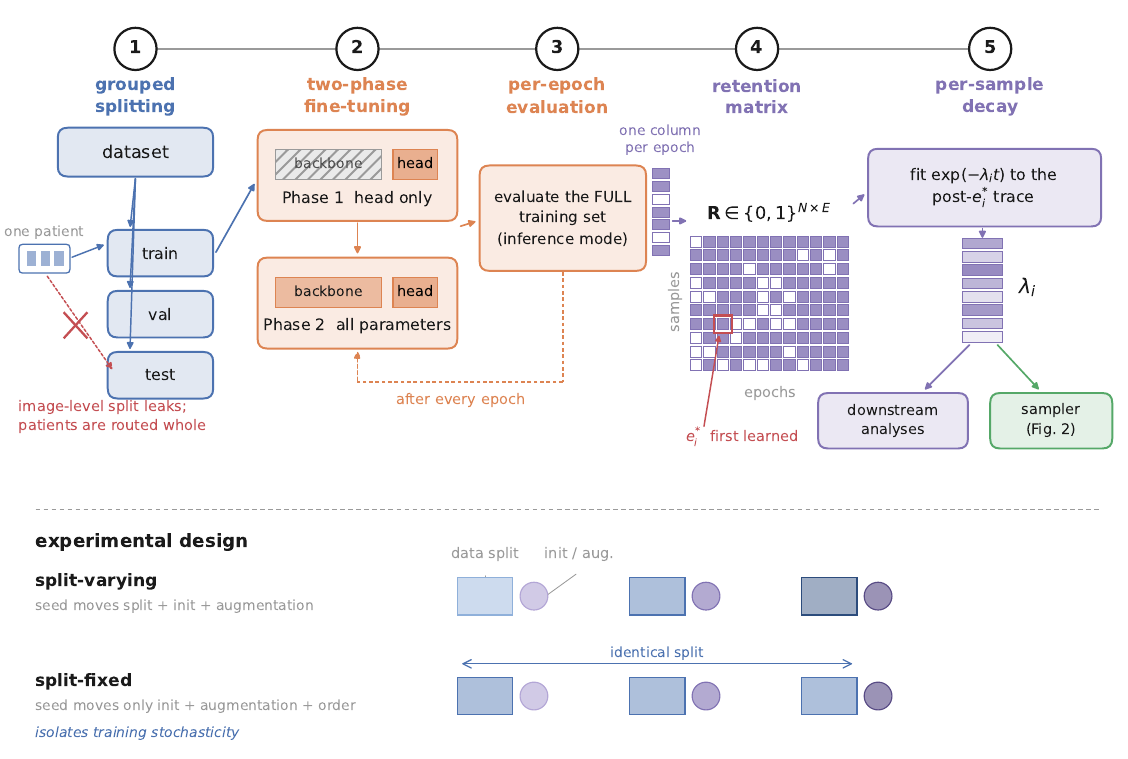}
  \caption{Pipeline and experimental design. Stage~1 splits the data by patient so that no patient contributes images to more than one partition; the struck-through route is the image-level split that leaks. Stages~2 and~3 run two-phase fine-tuning and evaluate the full training set in inference mode after every epoch. Stage~4 stacks those evaluations into the retention matrix $\mathbf{R}$, with $e^{*}_{i}$ marking the epoch at which sample $i$ is first classified correctly. Stage~5 fits a per-sample decay constant $\lambda_i$, which feeds the analyses and the sampler. The lower panel contrasts the two designs: under split-varying the seed changes the split as well as initialisation and augmentation, while under split-fixed one split is shared across all runs so that only training stochasticity varies.}
  \label{fig:framework}
\end{figure*}

We record whether each training sample is classified correctly at the end of every epoch, summarise each sample's resulting trace by a decay constant, and use those constants both as an object of study and as the input to a training sampler. The same pipeline runs under two designs, which we refer to throughout as \textit{split-varying}, where the random seed changes the data split along with initialisation and augmentation, and \textit{split-fixed}, where one split is shared across every run so that the seed changes only training stochasticity. Figure~\ref{fig:framework} shows the pipeline and both designs; Section~\ref{sec:setup} gives the seeds used for each.

\subsection{Per-sample retention tracking}\label{sec:tracking}

At the end of every training epoch we evaluate the entire training set under inference mode, with no augmentation and no gradients, and record a binary correctness indicator per sample. This gives a retention matrix $\mathbf{R} \in \{0,1\}^{N \times E}$ where $R_{i,e} = 1$ if sample $i$ is classified correctly at epoch $e$. The same forward pass also records each sample's cross-entropy loss, margin, confidence, and predicted class, so any of these can be substituted for correctness later without retraining.

For each sample we record the first-learned epoch $e^{*}_i = \min\{e : R_{i,e} = 1\}$, the number of forgetting events, and the retention rate $\rho_i$, defined as the fraction of epochs after $e^{*}_i$ at which the sample remains correct.

\subsection{Decay fitting and its edge cases}\label{sec:decay}

Following the Ebbinghaus form~\cite{ebbinghaus1885}, we model retention after first learning as an exponential in \eqref{eq:decay}.

\begin{equation}\label{eq:decay}
  P(\text{retained at time } t) = \exp(-\lambda_i \cdot t),
\end{equation}

where $t$ counts epochs since $e^{*}_i$ and $\lambda_i \geq 0$. Larger $\lambda_i$ means faster forgetting. Fitting uses nonlinear least squares~\cite{virtanen2020scipy} with $\lambda_i$ bounded to $[0, 10]$. That ceiling is never reached. The largest fitted value in any of the four configurations is $9.58$, and no fitted sample sits at the bound, so nothing here depends on where it was set.

Four cases arise, and separating them matters more than it first appears. A sample that is correct at every epoch after $e^{*}_i$ has a constant trace and exhibits no decay; it is assigned $\lambda_i = 0$ directly and is never passed to the optimiser, because fitting a curve to a constant target is undefined and the solver's stopping point is then determined by trace length rather than by the data. A sample never classified correctly has no trace to fit. A sample first learned within two epochs of the end has too few observations. Everything else is fitted. Each sample carries its category explicitly, and Section~\ref{sec:setup} states how the first three are handled in each analysis rather than folding them into a single imputed value.

Fit quality is reported only over fitted, non-constant traces. Including constant traces makes the average a count of samples that never forgot rather than a measure of how well the curve describes forgetting. We report $R^2$ on that restricted set alongside mean absolute error, which does not inherit the same degeneracy.

A monotone curve cannot follow a trace that is forgotten and then relearned (Section~\ref{sec:bg_forgetting}), so we fit two further families for comparison: an exponential with a non-zero asymptote, $c + (1-c)\exp(-\lambda_i t)$, and a power law, $(1+t)^{-a}$. We also compute transition rates directly from the trace, namely the probability of forgetting given currently correct and the probability of relearning given currently incorrect. These make no monotonicity assumption at all.

\subsection{What we compute from the decay constants}\label{sec:analysis}

Five quantities are computed from $\lambda$, at the sample, architecture, and class level.

\begin{enumerate}
  \item \textbf{Run-to-run reproducibility.} Spearman rank correlation of $\lambda$ between two runs. Samples are matched by image identity, not by position in the training array, because a different seed produces a different split and positional matching would compare unrelated images.
  \item \textbf{Within versus between architecture.} The same correlation computed between two runs of one architecture and between two architectures, with the split held fixed so the comparison isolates the architecture.
  \item \textbf{Cross-architecture set overlap.} Jaccard similarity of the top-10\% highest-$\lambda$ samples, reported against the chance baseline of Section~\ref{sec:bg_stats} and a permutation interval.
  \item \textbf{Class-level structure.} Mean $\lambda$ per class, tested against class size by rank correlation rather than asserted from inspection.
  \item \textbf{Early loss as a predictor.} Spearman correlation between a sample's Phase~1 loss and its fitted $\lambda$, computed both within a run and across runs, where the loss comes from one run and the decay constant from another.
\end{enumerate}

\subsection{Spaced repetition sampler}\label{sec:sampler}

The decay constants translate into a training sampler in the manner of spaced repetition systems~\cite{leitner1972so,settles2016trainable}. At epoch $e$, sample $i$ has gone $g_i = e - e_i^{\text{last}}$ epochs since it last appeared in a batch and receives the urgency in \eqref{eq:urgency}, which is the model's estimate of the probability that the sample has been forgotten since it was last seen.

\begin{equation}\label{eq:urgency}
  u_i^{(e)} = 1 - \exp\bigl(-\lambda_i \cdot g_i\bigr).
\end{equation}

Sampling probabilities follow from a softmax with temperature $\tau$, given in \eqref{eq:prob}.

\begin{equation}\label{eq:prob}
  P(i) = \frac{\exp(u_i^{(e)} / \tau)}{\sum_j \exp(u_j^{(e)} / \tau)}.
\end{equation}

The temperature is not a free stylistic choice, and \eqref{eq:bound} is the reason. Since $u_i^{(e)} \in [0,1)$ by construction, the ratio between the largest and smallest sampling probability is bounded above regardless of how the decay constants are distributed:

\begin{equation}\label{eq:bound}
  \frac{\max_i P(i)}{\min_i P(i)} \;=\; \exp\!\left(\frac{\max_i u_i - \min_i u_i}{\tau}\right) \;\leq\; \exp(1/\tau).
\end{equation}

\begin{figure*}[!h]
  \centering
  \includegraphics[width=\linewidth]{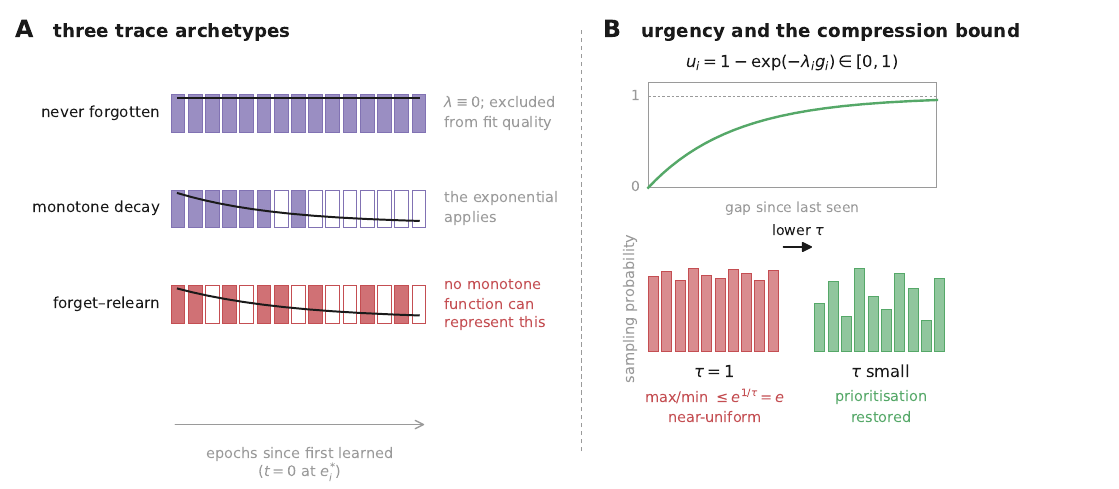}
  \caption{What $\lambda$ measures and where it breaks down. Panel A shows the three trace archetypes with the fitted exponential drawn over each: a never-forgotten trace carries no decay and is excluded from fit quality, a monotone trace is the case the model was designed for, and a forget-relearn trace cannot be represented by any monotonically decreasing function. The same three shapes appear as real traces with fitted $\lambda$ in Figure~\ref{fig:archetypes}B. Panel B shows that urgency is bounded in $[0,1)$, so the softmax in \eqref{eq:prob} caps the achievable spread of sampling probabilities at $\exp(1/\tau)$; at $\tau = 1$ the arms are near-uniform, and lowering $\tau$ restores prioritisation.}
  \label{fig:mechanism}
\end{figure*}

At $\tau = 1$ that ceiling is $e \approx 2.72$, so the sampler cannot depart far from uniform even when $\lambda$ varies over four orders of magnitude. Any evaluation at a single temperature therefore risks testing the temperature rather than the schedule, and we sweep $\tau$ accordingly. Figure~\ref{fig:mechanism} shows both this bound and the trace geometry that limits the decay model.

We also evaluate an online variant that discards the precomputed constants and re-estimates $\lambda_i$ during Phase~2 from the running forget rate of each sample, which removes the assumption that a constant measured in one run transfers to another. Baselines and the sampling diagnostics used to check whether an arm prioritises at all are given in Section~\ref{sec:setup}.

\section{Experimental Setup}\label{sec:setup}

This section fixes what a reader would need to repeat the measurements: the two datasets and how each is partitioned, the fine-tuning protocol and the horizons it produces, the two experimental designs, the five sampling arms and their diagnostics, and the statistical protocol. Three choices carry most of the weight. OCTDL is split over patients rather than images. One design holds that split fixed across seeds, leaving training stochasticity as the only thing that moves. And every arm logs its realised sampling distribution, which is what separates an arm that failed to help from one that never prioritised at all.

\subsection{Datasets}\label{sec:setup_data}

The two datasets stress different sources of difficulty. OCTDL~\cite{kulyabin2024octdl} is small and severely imbalanced; CUB-200-2011~\cite{wah2011caltech} is larger, balanced, and fine-grained. Table~\ref{tab:datasets} summarises both, including the unit at which each is split.

\begin{table}[!h]
\centering
\caption{Datasets and splitting. The split unit differs because OCT images from one patient are not independent, whereas CUB photographs are.}\label{tab:datasets}
\resizebox{\columnwidth}{!}{
\begin{tabular}{|l|c|c|c|c|c|}
\hline
\textbf{Dataset} & \textbf{Images} & \textbf{Classes} & \textbf{Imbalance} & \textbf{Split unit} & \textbf{Difficulty driver} \\
\hline
OCTDL   & 2,064  & 7   & 56:1  & patient (820) & class size \\
\hline
CUB-200 & 11,788 & 200 & 1.5:1 & image & visual similarity \\
\hline
\end{tabular}
}
\end{table}

\subsection{Patient-grouped splitting and what image-level splitting costs}\label{sec:setup_split}

OCTDL is split 70/15/15 over patients rather than over images, stratified by each patient's class, which yields 573 training, 123 validation, and 124 test patients. Grouping uses the numeric patient identifier alone rather than a class-patient pair, because one identifier in OCTDL appears under two diagnoses and a class-qualified key would separate that patient's images across partitions. Disjointness is asserted at run time rather than assumed.

The more common image-level split is easy to reproduce by accident, so its cost is set out here. Splitting the same data at the image level places between 75\% and 76\% of test patients in the training set as well, and 76\% to 78\% of individual test images share a patient with a training image. Section~\ref{sec:res_char} reports what this does to the headline metrics.

Patient counts are therefore fixed across seeds, while image counts vary by a few percent because patients contribute different numbers of images: seed 42 yields 1,428 training, 310 validation, and 326 test images, and the other two split-varying seeds yield 1,474 and 1,437 training images. CUB uses the official train/test partition with 15\% of the official training set held out for validation, giving 5,094 training, 900 validation, and 5,794 test images. Each photograph is independent, so no grouping applies.

\subsection{Training protocol}\label{sec:setup_train}

Both backbones are initialised from ImageNet-1K weights through \texttt{timm}~\cite{rw2019timm}. Phase~1 freezes the backbone and trains only the classification head for 5 epochs with AdamW at a learning rate of $10^{-3}$. Retention tracking runs from the first epoch, so Phase~1 is recorded like any other epoch, and the Phase~1 loss used in Section~\ref{sec:res_predictor} is the value from the end-of-epoch evaluation pass of Section~\ref{sec:tracking} at epoch 5, measured on unaugmented images in inference mode rather than taken from a training batch. Phase~2 unfreezes all parameters and continues for up to 45 further epochs at $10^{-4}$ with cosine annealing to $10^{-6}$, weight decay $10^{-4}$ throughout, and early stopping on validation loss with patience 10. Training augmentation is \texttt{RandomResizedCrop}(224), horizontal flip, and \texttt{ColorJitter}; evaluation uses \texttt{Resize}(256) and \texttt{CenterCrop}(224). Batch size is 32 except for DeiT-Small on CUB, which uses 16 to fit in memory. On OCTDL the retention runs draw batches with inverse-frequency class weighting, so expected exposure is roughly equal across classes despite the imbalance; CUB-200 is already balanced and is drawn by shuffling. Section~\ref{sec:res_class} says what that weighting does to the class-level analysis.

Early stopping does not always fire, and the resulting horizons differ between backbones and somewhat between designs. Under split-varying, ResNet-18 runs a mean of 46 epochs on OCTDL (range 39 to 50) and reaches the 50-epoch ceiling on every CUB seed, so its CUB runs are budget-limited rather than converged, while DeiT-Small stops after a mean of 19 epochs on OCTDL (range 17 to 22) and a flat 20 on CUB. Under split-fixed the spread is wider: ResNet-18 averages 40 epochs on OCTDL (range 33 to 50) and DeiT-Small 22 (range 17 to 33). On CUB-200, ResNet-18 reaches the 50-epoch ceiling on every split-fixed seed as well, while DeiT-Small stops after a mean of 20.5 epochs (range 19 to 23). Section~\ref{sec:res_arch} uses the split-fixed figures. Because a longer run offers more opportunities to record a forgetting event, Section~\ref{sec:res_char} repeats the fit-quality analysis with every run truncated to a common horizon.

Retention tracking itself is unaffected by batch size and augmentation, since it runs in inference mode on unaugmented images.

\subsection{The two designs}\label{sec:setup_design}

Under \textit{split-varying} the seed controls the data split, weight initialisation, and augmentation order together, which is the usual practice and the one an earlier version of this study followed. We refer to that earlier version as v1 below, where contrasting it with the present analysis makes a methodological point; it is superseded in full by this paper. Three seeds are used: 42, 99, and 2026. Two runs under this design see partially different training sets, so any comparison between them must match samples by image identity.

Under \textit{split-fixed} one split is generated once and reused, and the seed controls only initialisation, augmentation, and batch order. Ten seeds are used: 42, 99, 2026, 7, 13, 21, 55, 101, 202, and 303. This is the design that separates training stochasticity from data composition, and it is the one the reproducibility results in Section~\ref{sec:res_stability} rest on. All four dataset and backbone combinations are run under both designs.

\subsection{Sampling arms and diagnostics}\label{sec:setup_sampling}

Phase~1 is identical across arms. Arms differ only in how Phase~2 draws its batches, and every arm draws the same number of samples with replacement so that per-epoch exposure is matched by construction. Five arms are compared:

\begin{enumerate}
  \item \textbf{Random}, the baseline used previously: inverse-frequency class weighting on OCTDL, uniform on the balanced CUB.
  \item \textbf{Uniform}, drawn uniformly with replacement on both datasets. Comparing this against Random isolates class balancing, and against the remaining arms isolates per-sample scheduling.
  \item \textbf{Curriculum}~\cite{bengio2009curriculum}, ranked by Phase~1 loss with easy samples favoured early and the weighting relaxing linearly toward uniform.
  \item \textbf{Anti-curriculum}, the same schedule reversed.
  \item \textbf{Spaced repetition}, sampling by the urgency of \eqref{eq:urgency}.
\end{enumerate}

Because \eqref{eq:bound} caps how far the spaced-repetition arm can depart from uniform at any given temperature, we sweep $\tau \in \{1.0, 0.7, 0.5, 0.4, 0.3\}$ on OCTDL. These values were chosen by evaluating \eqref{eq:bound} against the observed decay constants before running anything: they span realised probability ratios of roughly $2.7\times$ to $28\times$. Temperatures below $0.3$ were excluded because they concentrate sampling so tightly that most of the training set goes unseen in an epoch. The online variant of Section~\ref{sec:sampler} is run over the same temperatures below $1.0$.

Every arm logs its realised sampling distribution at each epoch: the Gini coefficient and normalised entropy of the sampling probabilities, the fraction of the training set drawn at least once, and the share of draws going to the most urgent 5\% of samples. A run is flagged automatically if its probability ratio falls below $4\times$, which means it is not prioritising, or if it draws fewer than 40\% of unique samples per epoch, which means it is starving most of the data. These diagnostics answer whether an arm did anything, independently of whether it helped.

\subsection{Statistical analysis}\label{sec:setup_stats}

Aggregates over seeds are reported as mean $\pm$ sample standard deviation ($\mathrm{ddof}=1$). Comparisons between sampling arms are paired by seed and tested with a paired $t$-test and a Wilcoxon signed-rank test, with the bootstrap interval of the paired difference and Hedges' $g$ reported alongside, since three seeds give little power on their own. We report the Wilcoxon test only for completeness: at $n=3$ its smallest attainable two-sided $p$-value is $0.25$, so it cannot reach the $0.05$ threshold at any effect size. The $t$-test, the bootstrap interval and Hedges' $g$ carry the inference. 

Where two sampling arms share a test set we also run McNemar's test~\cite{mcnemar1947note} at the level of individual test samples, which is the higher-powered unit. The split comparison of Section~\ref{sec:res_char} is not of this kind. On OCTDL its two pipelines evaluate on different test partitions, so no sample-level pairing exists. On CUB-200 they do share a test set, and McNemar would apply in principle, but the earlier pipeline recorded only aggregate test metrics rather than per-image predictions, so no contingency table can be built from what was kept. Table~\ref{tab:leakcost} states the test used instead.

Every family of comparisons is corrected with Benjamini-Hochberg~\cite{benjamini1995controlling}, and both raw and corrected values are given. Class-level rank correlations are tested by permutation rather than by the asymptotic approximation, exhaustively over all $5{,}040$ orderings for OCTDL's seven classes and over 20,000 resamples for CUB-200's two hundred. Jaccard overlaps are reported against the analytic chance baseline of Section~\ref{sec:bg_stats} together with a permutation interval from 2,000 resamples; bootstrap intervals also use 2,000 resamples.

\subsection{Implementation and reproducibility}\label{sec:setup_impl}

All runs use a single NVIDIA Tesla T4. Every run under the two designs of Section~\ref{sec:setup_design} records its full configuration, library versions, and seeds alongside its outputs, and separately saves per-epoch per-sample correctness, loss, margin, confidence, and prediction; the split files with group identifiers; test-set predictions and probabilities; and model weights. The sampling arms of Section~\ref{sec:setup_sampling} are logged more coarsely, saving strategy, temperature and seed; per-epoch rather than per-sample loss and accuracy; the realised sampling distribution; and the same test-set predictions, probabilities and weights. Those runs carry no library-version record. No reported comparison sets a sampling-arm figure against a differently versioned pipeline, so that gap does not bear on any result here.

Every statistic in Section~\ref{sec:results} is recomputed from one of these two records by a separate CPU-only analysis step, so our own re-derivation required no further GPU time. Table~\ref{tab:leakcost} is the exception: its image-level column comes from the earlier pipeline described in Section~\ref{sec:setup_stats}, and its $\Delta$, SE, $p$ and $g$ columns compare against that legacy figure rather than deriving from these records alone.

\section{Results and Discussion}\label{sec:results}

We report the retention analysis first (Sections~\ref{sec:res_char}--\ref{sec:res_predictor}), then the sampler (Section~\ref{sec:res_sampler}). Table~\ref{tab:findings} states each question and the answer the data gives; the sections that follow supply the evidence. Four of the seven answers are negative, and one of those negatives is the sampler this work set out to build.

\begin{table*}[!h]
\centering
\caption{The seven questions this study asks, and what the measurements return. The final row compares image-level against patient-grouped splitting on OCTDL, with CUB-200 as a control whose grouping rule does not change. Correlations are Spearman $\rho$, except the exponential-fit row, which reports $R^2$. The architecture and reproducibility rows use the split-fixed design over ten seeds; every other row uses split-varying over three seeds. The architecture gap is the smaller of the two within-architecture correlations minus the between-architecture value, which is the more conservative of the two possible comparisons. Each cell gives the ResNet-18 and DeiT-Small values or their range.}\label{tab:findings}
\resizebox{\textwidth}{!}{
\begin{tabular}{|l|c|c|c|}
\hline
\textbf{Question} & \textbf{OCTDL} & \textbf{CUB-200} & \textbf{Answer} \\
\hline
Do the two architectures forget different samples? & $+0.145$ & $+0.173$ & yes \\
\hline
Does the exponential describe per-sample retention? & $-0.050$ / $-0.004$ & $-0.074$ / $-0.080$ & no \\
\hline
Is per-sample forgetting reproducible across runs? & $0.45$--$0.59$ & $0.37$--$0.44$ & yes \\
\hline
Does class size predict mean $\lambda$? & $0.00$ / $-0.18$ & $+0.11$ / $+0.08$ & no \\
\hline
Does Phase~1 loss predict $\lambda$? & $0.39$--$0.43$ & $0.29$--$0.39$ & yes \\
\hline
Does $\lambda$-driven sampling beat uniform? & \multicolumn{2}{c|}{0 of 48 tests survive BH-FDR} & no \\
\hline
Does patient-grouped splitting change the metrics? & $+0.3$ / $+0.5$ pts & $-0.4$ / $-0.2$ pts & no \\
\hline
\end{tabular}
}
\end{table*}

\subsection{Forgetting curve characterisation}\label{sec:res_char}

\begin{figure*}[!h]
  \centering
  \includegraphics[width=\textwidth]{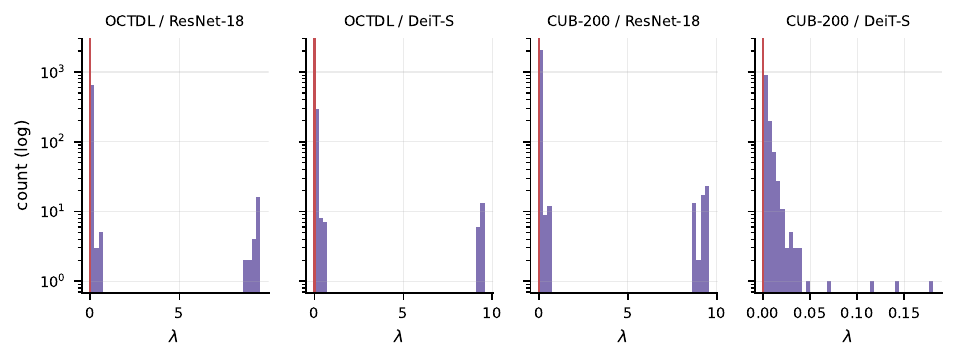}
  \caption{Distribution of fitted decay constants. The spike at $\lambda = 0$ is the constant traces, which make up the majority of samples in every configuration.}
  \label{fig:lambda_hist}
\end{figure*}

\begin{figure*}[!h]
  \centering
  \includegraphics[width=\textwidth]{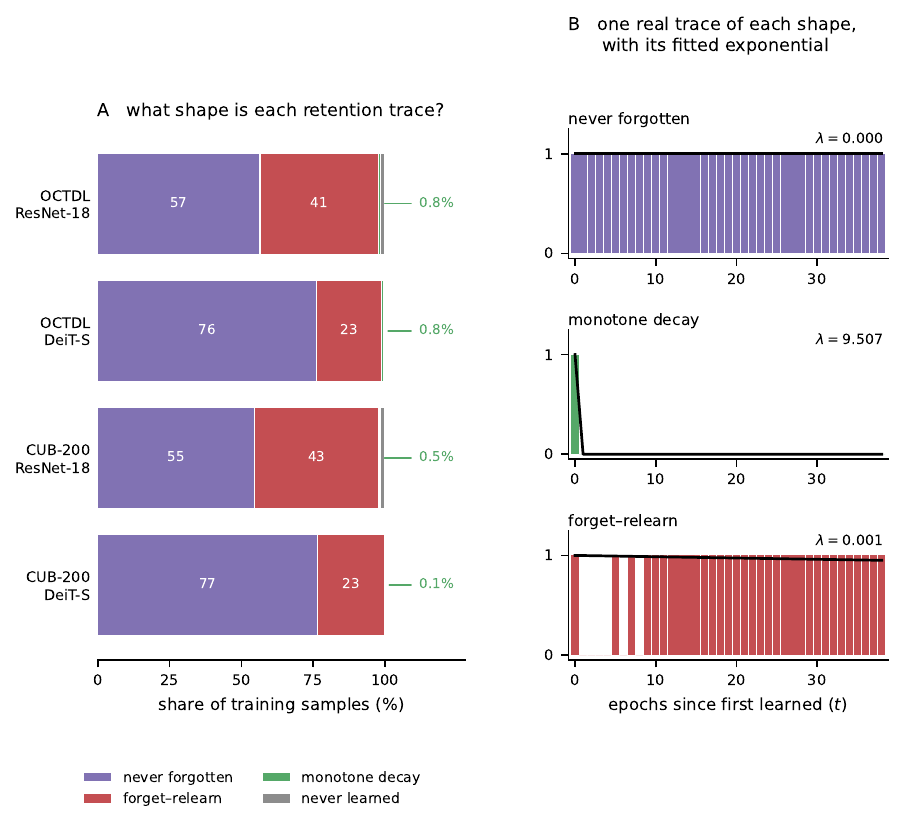}
  \caption{The four trace shapes, and what the exponential makes of them. Panel A gives the share of training samples in each shape, averaged over the three split-varying seeds. Monotone decay, the shape \eqref{eq:decay} is built for, accounts for between $0.1\%$ and $0.8\%$ of samples, the lowest figure resting on nine traces in total. Panel B shows one real trace of each of the three shapes worth fitting, introduced schematically in Figure~\ref{fig:mechanism}A, from OCTDL with ResNet-18; a never-learned trace is flat at zero throughout and is omitted. The twelve monotone traces in that run divide by how long they last: nine are lost one epoch after being learned and pin $\lambda$ near its upper bound, as the exemplar does, while three last two epochs and fit $\lambda \approx 0.64$. Neither value describes a decay. The forget-relearn exemplar draws $\lambda \approx 0$ despite three forgetting events; it was chosen to show several recoveries, and most forget-relearn traces record only one.}
  \label{fig:archetypes}
\end{figure*}

The decay constants pile up at zero. On OCTDL, 56.5\% of ResNet-18 traces and 76.2\% of DeiT-Small traces record no forgetting event at all once the sample is first classified correctly; on CUB-200 the figures are 54.6\% and 76.6\%. Figure~\ref{fig:lambda_hist} shows the resulting distributions, and Figure~\ref{fig:archetypes} shows what the underlying traces look like.

Sorting the traces by shape sharpens this. A trace is never learned, constant once it is first correct, monotone if it never recovers after its first lapse, or forget-relearn if it does. Figure~\ref{fig:archetypes} gives the split: monotone decay, the one shape \eqref{eq:decay} was written for, accounts for $0.8\%$ of samples on OCTDL under both backbones, $0.5\%$ on CUB-200 with ResNet-18 and $0.1\%$ with DeiT-Small, and never-learned traces for a further $0.0\%$ to $1.6\%$. The rest divide between constant traces, which carry no decay at all, and forget-relearn traces, which no monotone function can follow. Fitting a decay curve here is fitting a shape the data almost never takes.

Those constant traces decide how the fit quality reads. A trace that is 1 at every epoch after $e^*_i$ has zero variance, so $R^2$ has a zero denominator and MAE is exactly zero. Scoring them as perfect fits turns the reported $R^2$ into a count of how many samples were never forgotten. The uncorrected column of Table~\ref{tab:fitquality} shows what that produces: 0.552 and 0.765 on OCTDL, 0.522 and 0.747 on CUB-200, values that track the constant-trace percentages to within roughly two points in all four cases, between 0.2 and 2.4. What looked like a finding that DeiT retention is better described by exponential decay is a restatement of the fact that DeiT forgets fewer samples.

\begin{figure*}[!h]
  \centering
  \includegraphics[width=\textwidth]{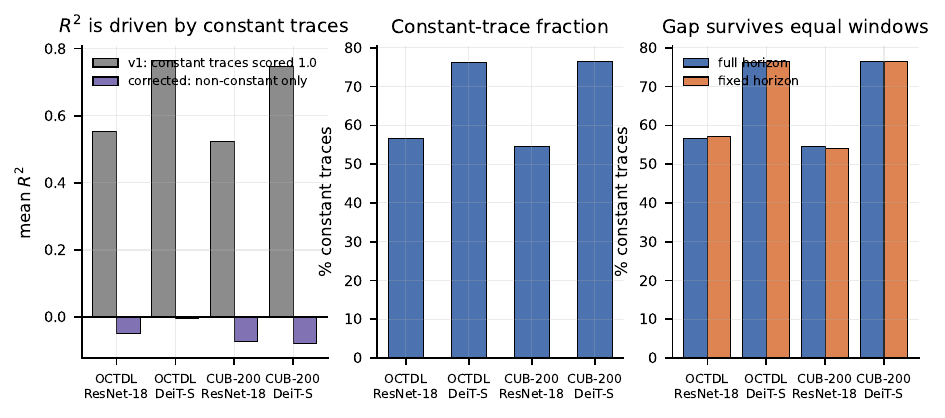}
  \caption{Per-sample $R^2$ scored only on traces with at least one forgetting event. Mass below zero indicates samples the exponential fits worse than their own mean.}
  \label{fig:fitquality}
\end{figure*}

Restricting the score to traces with at least one forgetting event reverses it. Corrected $R^2$ is $-0.050$ and $-0.004$ on OCTDL, $-0.074$ and $-0.080$ on CUB-200. A negative $R^2$ means the fitted exponential predicts the trace worse than that sample's own mean would. Figure~\ref{fig:fitquality} shows the per-sample distributions behind these averages. The honest reading is that the exponential does not describe per-sample retention for either architecture. What $\lambda$ actually separates is whether a sample is ever forgotten, not how quickly it decays once it is.

A different monotone form does not rescue it. Section~\ref{sec:decay} fits two further families to the same traces: an exponential with a free asymptote, which is allowed to level off above zero, and a power law, which falls away more slowly at first. Neither describes the data either (Table~\ref{tab:altmodels}). The power law stays negative in three of the four configurations. The asymptotic exponential turns positive in all four, but only just, between $+0.012$ and $+0.087$, and it buys that with a second free parameter, so some improvement was guaranteed before any data was seen. Mean absolute error tells the plainer story: both alternatives are worse than the plain exponential in every configuration, by $0.02$ to $0.04$. The extra parameter lets the asymptotic model sit at roughly the right average level without following the trace any better. What fails here is not the choice of monotone function but monotonicity itself.

One alternative explanation deserves a test. DeiT converges in 17 to 22 epochs while ResNet runs for 39 to 50, so DeiT traces have less time in which a forgetting event could occur, and the constant-trace gap might be an artifact of the observation window. Truncating every run to the shortest horizon in its dataset ($E = 17$ on OCTDL, $E = 20$ on CUB-200) leaves the gap intact: 57.1\% versus 76.5\% and 54.2\% versus 76.6\%, with corrected $R^2$ still at or near zero, negative in three configurations and $+0.003$ in the fourth (Table~\ref{tab:fitquality}, final column). The difference is a property of the architectures under matched exposure, not of how long we watched.

Image-level splitting, whose scale Section~\ref{sec:setup_split} sets out, moves the headline metrics far less than that scale suggests. Table~\ref{tab:leakcost} sets the patient-grouped runs against runs of the v1 pipeline on image-level splits of the same data, using the same pretrained initialisation and training hyperparameters. On OCTDL, accuracy under the leaky split is higher by 0.3 points for ResNet-18 and 0.5 for DeiT-Small, while macro F1 is lower under it for both. CUB-200 is split by image in both cases and so acts as a control. Its accuracy moves 0.2 to 0.4 points the other way, lower under the leaky split rather than higher, while its macro F1 moves by a similar amount in the same direction as OCTDL's. Macro F1 is therefore lower under the leaky split on both datasets, and only accuracy changes sign between them. No difference on either metric exceeds 1.3 times its own standard error, and no test approaches significance, the smallest Welch $p$ being 0.29.

That cuts against the usual expectation. Three quarters or more of the OCTDL test set shares a patient with training, and the aggregate metrics barely register it. Patient-grouped splitting is still the right choice, because the leak is real and a metric that happens not to move on one dataset guarantees nothing on another. On this dataset, though, the practice does not buy the correction it is usually assumed to.

Two limits on that reading deserve stating. The two split procedures do not merely resize the partitions, they reallocate the rare classes: RAO, with 15 to 19 training images depending on the split, contributes three test images under the image-level split and one under the patient-grouped split, and a class that small can move macro F1 by more than any difference in Table~\ref{tab:leakcost}. And the v1 runs predate the provenance logging described in Section~\ref{sec:setup_impl}, so their library versions cannot be confirmed identical to ours. We therefore read this comparison as indicative rather than as a controlled ablation.

Traces that do show forgetting are not decaying in the sense the model assumes. Estimating a two-state Markov chain on them gives a per-epoch forgetting probability of 0.020 to 0.051 against a relearning probability of 0.47 to 0.70. Forgetting is rare and recovery from it is close to a coin flip, which produces the forget-relearn traces in the third row of Figure~\ref{fig:archetypes}. No monotonically decreasing function can represent that shape, and the negative $R^2$ values are the arithmetic consequence.

\begin{table*}[!h]
\centering
\caption{Exponential fit quality, over three split-varying seeds. The uncorrected column scores constant traces as $R^2=1$, as is common practice; the corrected column restricts scoring to traces with at least one forgetting event. The final column repeats the corrected value with every run truncated to a common horizon.}\label{tab:fitquality}

\begin{tabular}{|l|c|c|c|c|c|}
\hline
\textbf{Dataset / backbone} & \textbf{\% constant} & \textbf{$R^2$ (uncorrected)} & \textbf{$R^2$ corrected} & \textbf{MAE} & \textbf{$R^2$ fixed horizon} \\
\hline
OCTDL / ResNet-18 & 56.5\% & 0.552 & $-0.050$ & 0.095 & $-0.112$ \\
\hline
OCTDL / DeiT-S & 76.2\% & 0.765 & $-0.004$ & 0.190 & $+0.003$ \\
\hline
CUB-200 / ResNet-18 & 54.6\% & 0.522 & $-0.074$ & 0.100 & $-0.080$ \\
\hline
CUB-200 / DeiT-S & 76.6\% & 0.747 & $-0.080$ & 0.122 & $-0.080$ \\
\hline
\end{tabular}

\end{table*}

\begin{table*}[!h]
\centering
\caption{Three monotone families fitted to the same traces, scored over fitted non-constant traces only, mean over three split-varying seeds. The asymptotic exponential carries a second free parameter, so its $R^2$ is not directly comparable with the other two. No family reaches $R^2 = 0.09$, and both alternatives are worse than the plain exponential on mean absolute error.}\label{tab:altmodels}

\begin{tabular}{|l|c|c|c|c|c|c|}
\hline
& \multicolumn{3}{c|}{\textbf{$R^2$}} & \multicolumn{3}{c|}{\textbf{MAE}} \\
\cline{2-7}
\textbf{Dataset / backbone} & \textbf{exp.} & \textbf{+ asym.} & \textbf{power} & \textbf{exp.} & \textbf{+ asym.} & \textbf{power} \\
\hline
OCTDL / ResNet-18 & $-0.050$ & $+0.033$ & $-0.027$ & 0.095 & 0.133 & 0.112 \\
\hline
OCTDL / DeiT-S & $-0.004$ & $+0.087$ & $+0.022$ & 0.190 & 0.216 & 0.209 \\
\hline
CUB-200 / ResNet-18 & $-0.074$ & $+0.021$ & $-0.043$ & 0.100 & 0.136 & 0.118 \\
\hline
CUB-200 / DeiT-S & $-0.080$ & $+0.012$ & $-0.056$ & 0.122 & 0.163 & 0.141 \\
\hline
\end{tabular}

\end{table*}

\begin{table*}[!h]
\centering
\caption{What image-level splitting costs on the headline metrics. OCTDL is split by image in the first column and by patient in the second; CUB-200 is split by image in both and acts as a control, since no grouping rule changes for it. The grouped values come from the retention runs of Section~\ref{sec:setup_design}, which are separate training runs from the sampling runs of Table~\ref{tab:main}. Mean $\pm$ sd over three seeds, $\mathrm{ddof}=1$. $\Delta$ is image-level minus grouped, from unrounded means. OCTDL's two pipelines evaluate on genuinely different test partitions; CUB-200's evaluate the same official test images under independently trained checkpoints. In both cases the per-seed values are treated as draws from independent runs rather than as paired observations, so SE is the two-sample standard error, $p$ comes from Welch's $t$-test, and $g$ is Hedges' $g$. No comparison approaches significance, and correcting these eight tests for multiplicity would only raise the $p$-values further.}\label{tab:leakcost}

\begin{tabular}{|l|l|l|c|c|c|c|c|c|}
\hline
\textbf{Dataset} & \textbf{Backbone} & \textbf{Metric} & \textbf{image-level} & \textbf{grouped} & \textbf{$\Delta$} & \textbf{SE} & \textbf{$p$} & \textbf{$g$} \\
\hline
OCTDL & ResNet-18 & Accuracy & 0.916$\pm$0.012 & 0.914$\pm$0.027 & $+0.0028$ & 0.0171 & 0.88 & $+0.11$ \\
\hline
OCTDL & ResNet-18 & Macro F1 & 0.851$\pm$0.012 & 0.856$\pm$0.044 & $-0.0053$ & 0.0265 & 0.86 & $-0.13$ \\
\hline
OCTDL & DeiT-S & Accuracy & 0.909$\pm$0.012 & 0.904$\pm$0.012 & $+0.0052$ & 0.0100 & 0.63 & $+0.34$ \\
\hline
OCTDL & DeiT-S & Macro F1 & 0.825$\pm$0.049 & 0.833$\pm$0.026 & $-0.0083$ & 0.0322 & 0.81 & $-0.17$ \\
\hline
CUB-200 & ResNet-18 & Accuracy & 0.693$\pm$0.005 & 0.697$\pm$0.003 & $-0.0039$ & 0.0032 & 0.31 & $-0.79$ \\
\hline
CUB-200 & ResNet-18 & Macro F1 & 0.692$\pm$0.005 & 0.696$\pm$0.003 & $-0.0043$ & 0.0034 & 0.29 & $-0.83$ \\
\hline
CUB-200 & DeiT-S & Accuracy & 0.775$\pm$0.002 & 0.777$\pm$0.003 & $-0.0021$ & 0.0020 & 0.37 & $-0.67$ \\
\hline
CUB-200 & DeiT-S & Macro F1 & 0.773$\pm$0.003 & 0.775$\pm$0.003 & $-0.0022$ & 0.0024 & 0.42 & $-0.59$ \\
\hline
\end{tabular}

\end{table*}

\subsection{Architecture-dependent forgetting}\label{sec:res_arch}

\begin{figure}[!h]
  \centering
  \includegraphics[width=\linewidth]{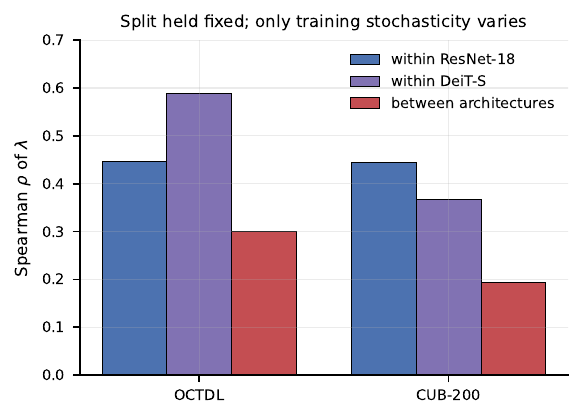}
  \caption{Within- versus between-architecture agreement on $\lambda$ with the split held fixed. The separation is the architecture effect.}
  \label{fig:withinbetween}
\end{figure}

Holding the split fixed lets us ask whether ResNet-18 and DeiT-Small forget different samples, or whether they forget the same samples and the apparent difference is training noise. Two runs of the same architecture agree on $\lambda$ at $\rho = 0.45$ and $0.59$ on OCTDL and $0.44$ and $0.37$ on CUB-200. Two runs of different architectures agree at $0.30$ and $0.19$. The within-architecture value exceeds the between-architecture value on both datasets, by 0.145 and 0.173 (Table~\ref{tab:arch}, Figure~\ref{fig:withinbetween}). Since the two comparisons share a split, the same augmentation policy and the same optimiser, what separates the two backbones is the architecture and whatever else differs with it. On OCTDL that is the training horizon alone; on CUB-200 it is the horizon and the batch size. Under the split-fixed design this analysis uses, DeiT-Small runs for 17 to 33 epochs on OCTDL and ResNet-18 for 33 to 50. On CUB the disparity is larger still: ResNet-18 spends the full 50-epoch budget on every seed while DeiT-Small stops after a mean of 20.5, and the two also differ in batch size there. A DeiT trace therefore offers roughly half as many post-$e^{*}_i$ observations for the fit, and its $\lambda$ carries correspondingly more estimation noise. The two ranges overlap, so horizon is not perfectly confounded with architecture, but that is not enough on its own. Repeating the fixed-horizon control of Section~\ref{sec:res_char} for this correlation settles the horizon question. Truncating every split-fixed run to the shortest in its dataset, $E = 17$ on OCTDL and $E = 19$ on CUB-200, lowers all three correlations, which is what fitting each $\lambda$ to fewer observations should do and costs ResNet-18 the most since it gives up the largest share of its epochs. The gap survives: $+0.127$ on OCTDL and $+0.194$ on CUB-200, against $+0.145$ and $+0.173$ at full length (Table~\ref{tab:arch}). On OCTDL, where horizon was the only other difference, that leaves the architecture as the remaining explanation. On CUB-200 the smaller batch size used for DeiT-Small is still uncontrolled and we cannot separate it without retraining, so there the control narrows the explanation without closing it.

The gap is small in absolute terms. Between-architecture agreement of 0.19 to 0.30 is well above zero, so the two backbones are far from forgetting disjoint sets of samples. Put another way, training stochasticity alone leaves two runs disagreeing at a rank correlation of $0.37$ to $0.59$, and letting the architecture differ as well drops that to $0.19$ to $0.30$.

\begin{figure*}[!h]
  \centering
  \includegraphics[width=\textwidth]{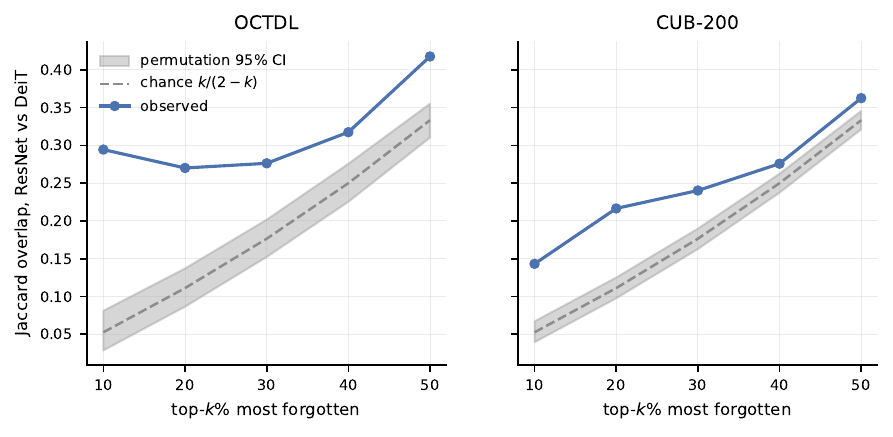}
  \caption{Top-$k\%$ Jaccard overlap between architectures against the chance baseline $k/(2-k)$. Excess over chance is largest in the tail.}
  \label{fig:jaccard}
\end{figure*}

Ranking overlap tells the same story once chance is accounted for. Taking the top-$k\%$ highest-$\lambda$ samples from each architecture and computing the Jaccard index gives 0.294 on OCTDL and 0.143 on CUB-200 at $k = 10$. Both look small until compared against the chance value $k/(2-k) = 0.053$, which makes them 5.6$\times$ and 2.7$\times$ chance and puts both outside the permutation interval (Table~\ref{tab:jaccard}, Figure~\ref{fig:jaccard}). The ratio falls monotonically as $k$ grows, reaching 1.25$\times$ and 1.09$\times$ at $k = 50$, which is what one expects when the agreement is concentrated in the extreme tail. Read without the chance baseline, the raw overlaps invert this reading: the smaller raw number at $k = 10$ is in fact the stronger evidence of shared structure.

\begin{table*}[!h]
\centering
\caption{Within- versus between-architecture agreement of $\lambda$, split held fixed so that only training stochasticity and the architecture differ. The final column takes the smaller of the two within-architecture correlations, the more conservative comparison. The lower block repeats the measurement with every run truncated to the shortest horizon in its dataset, so that both backbones contribute the same number of epochs to each fit.}\label{tab:arch}

\begin{tabular}{|l|c|c|c|c|c|}
\hline
\textbf{Dataset} & \textbf{Horizon} & \textbf{within ResNet-18} & \textbf{within DeiT-S} & \textbf{between} & \textbf{within $-$ between} \\
\hline
OCTDL & full & $+0.446$ & $+0.588$ & $+0.300$ & $+0.145$ \\
\hline
CUB-200 & full & $+0.444$ & $+0.366$ & $+0.193$ & $+0.173$ \\
\hline
OCTDL & $E=17$ & $+0.390$ & $+0.561$ & $+0.263$ & $+0.127$ \\
\hline
CUB-200 & $E=19$ & $+0.381$ & $+0.366$ & $+0.172$ & $+0.194$ \\
\hline
\end{tabular}

\end{table*}

\begin{table*}[!h]
\centering
\caption{Overlap of the top-$k\%$ highest-$\lambda$ samples between architectures, against the chance value $k/(2-k)$ and a permutation interval.}\label{tab:jaccard}

\begin{tabular}{|l|c|c|c|c|c|}
\hline
\textbf{Dataset} & \textbf{$k$} & \textbf{observed $J$} & \textbf{chance} & \textbf{ratio} & \textbf{permutation 95\% CI} \\
\hline
OCTDL & 10\% & 0.294 & 0.053 & 5.59$\times$ & [0.029, 0.081] \\
\hline
OCTDL & 20\% & 0.270 & 0.111 & 2.43$\times$ & [0.086, 0.137] \\
\hline
OCTDL & 30\% & 0.276 & 0.176 & 1.57$\times$ & [0.152, 0.202] \\
\hline
OCTDL & 40\% & 0.317 & 0.250 & 1.27$\times$ & [0.225, 0.276] \\
\hline
OCTDL & 50\% & 0.418 & 0.333 & 1.25$\times$ & [0.310, 0.355] \\
\hline
CUB-200 & 10\% & 0.143 & 0.053 & 2.72$\times$ & [0.039, 0.067] \\
\hline
CUB-200 & 20\% & 0.217 & 0.111 & 1.95$\times$ & [0.098, 0.125] \\
\hline
CUB-200 & 30\% & 0.240 & 0.176 & 1.36$\times$ & [0.163, 0.190] \\
\hline
CUB-200 & 40\% & 0.276 & 0.250 & 1.10$\times$ & [0.237, 0.262] \\
\hline
CUB-200 & 50\% & 0.363 & 0.333 & 1.09$\times$ & [0.321, 0.346] \\
\hline
\end{tabular}

\end{table*}

\subsection{Forgetting stability across seeds}\label{sec:res_stability}

Whether per-sample forgetting reproduces across runs turns entirely on how samples are matched between them, and this is where our earlier analysis went wrong.

Comparing per-sample statistics across two runs requires knowing which row of one run corresponds to which row of the other. Under a varying split the training sets are different subsets in a different order, so row $i$ of one run and row $i$ of another are generally different images. Two split-varying runs share 988 to 1{,}007 of their OCTDL training images and 4{,}328 to 4{,}336 of their CUB-200 images, between $69\%$ and $85\%$ of the smaller training set of the pair; under split-fixed the overlap is complete by construction. Correlating them by position measures nothing. Doing so here returns $\rho = 0.002$ to $0.137$, which is the near-zero value v1 reported and interpreted as evidence that per-sample forgetting is not reproducible.

\begin{figure*}[!h]
  \centering
  \includegraphics[width=\textwidth]{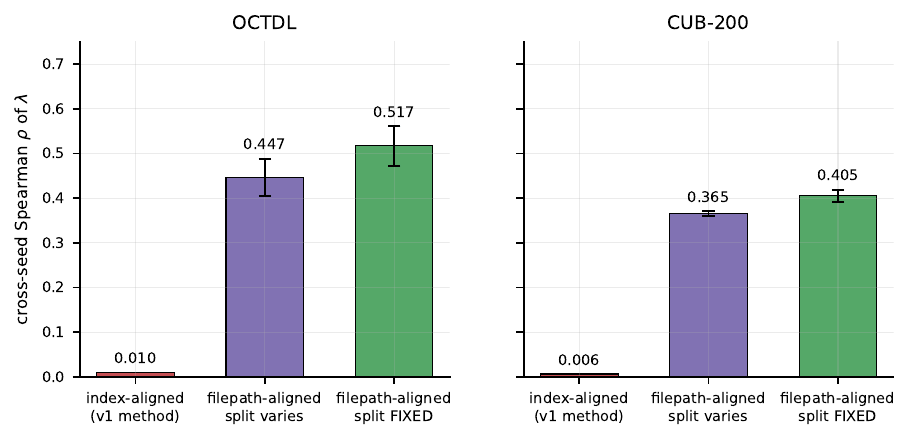}
  \caption{Run-to-run agreement on $\lambda$ under positional alignment and under a filepath join, for the same runs. The left column reproduces what positional alignment reports.}
  \label{fig:stability}
\end{figure*}

Joining on image filepath instead gives $\rho = 0.366$ and $0.527$ on OCTDL and $0.382$ and $0.347$ on CUB-200, over the same runs and the same fitted constants (Table~\ref{tab:stability}, Figure~\ref{fig:stability}). Holding the split fixed and averaging over all 45 seed pairs raises it to $0.446$, $0.588$, $0.444$ and $0.366$. Those 45 pairs come from 10 independent runs, not 45, so their spread understates the uncertainty. Resampling the runs rather than the pairs gives 95\% intervals of $[0.419, 0.475]$, $[0.559, 0.618]$, $[0.434, 0.453]$ and $[0.358, 0.373]$ for the four configurations, between about two and two and a half times the width a naive treatment of the pairs as independent would report, and all four sit well clear of zero. Pairs formed between two resampled copies of the same run are excluded, since a run correlates with itself perfectly and would inflate the interval. The same pattern holds for the other three per-sample statistics we track, with first-learned epoch the most reproducible of the four at up to $0.716$.

So per-sample forgetting is reproducible. Two independently seeded runs of the same architecture on the same split agree on which samples are fragile at $\rho \approx 0.37$ to $0.59$, and roughly a tenth of that correlation is attributable to the split rather than to training stochasticity. This is a moderate correlation, not a strong one, and it is measured on a $\lambda$ that Section~\ref{sec:res_char} showed to be a poor decay estimate. What is stable is the coarse ordering of samples by fragility.

The positional-alignment bug produced a clean, publishable negative result that survived our own review, and it failed silently: no exception, no missing data, no implausible value. Any per-sample analysis across runs with different data ordering needs an explicit identity join, and the cheapest check is that a run correlated against itself under the same code path returns exactly 1.

\begin{table*}[!h]
\centering
\caption{Run-to-run rank correlation of four per-sample statistics, matched by image identity. Under split-varying the seed also changes the split; under split-fixed it does not. Each cell is the mean over all seed pairs: 3 pairs from 3 runs under split-varying, 45 pairs from 10 runs under split-fixed. Pairs drawn from a shared pool of runs are not independent, so the number of pairs is not an effective sample size.}\label{tab:stability}
\resizebox{\textwidth}{!}{
\begin{tabular}{|l|c|c|c|c|c|c|c|c|}
\hline
& \multicolumn{4}{c|}{\textbf{split-varying (3 seeds)}} & \multicolumn{4}{c|}{\textbf{split-fixed (10 seeds)}} \\
\cline{2-9}
\textbf{Dataset / backbone} & $\lambda$ & \textbf{n forget} & \textbf{ret. rate} & \textbf{first lrn} & $\lambda$ & \textbf{n forget} & \textbf{ret. rate} & \textbf{first lrn} \\
\hline
OCTDL / ResNet-18 & $+0.366$ & $+0.309$ & $+0.405$ & $+0.378$ & $+0.446$ & $+0.385$ & $+0.475$ & $+0.412$ \\
\hline
OCTDL / DeiT-S & $+0.527$ & $+0.495$ & $+0.536$ & $+0.705$ & $+0.588$ & $+0.561$ & $+0.603$ & $+0.716$ \\
\hline
CUB-200 / ResNet-18 & $+0.382$ & $+0.310$ & $+0.419$ & $+0.478$ & $+0.444$ & $+0.367$ & $+0.471$ & $+0.519$ \\
\hline
CUB-200 / DeiT-S & $+0.347$ & $+0.352$ & $+0.370$ & $+0.558$ & $+0.366$ & $+0.367$ & $+0.387$ & $+0.562$ \\
\hline
\end{tabular}
}
\end{table*}

\subsection{Class-level forgetting patterns}\label{sec:res_class}

On OCTDL, RVO carries the highest mean decay constant under both backbones, 0.876 for ResNet-18 and 0.845 for DeiT-Small, and in each case the lowest never-forgotten rate of any class, 34.2\% for ResNet-18 and 42.2\% for DeiT-Small (Table~\ref{tab:octdl_class}). AMD, nearly twelve times larger at 858 training images, sits near the bottom at 0.084. That pairing invites the conclusion that small classes are forgotten more.

The full ranking does not support it. Across all seven classes the correlation between training-set size and mean $\lambda$ is $\rho = 0.000$ for ResNet-18 and $-0.179$ for DeiT-Small, with exact permutation $p$ of $1.000$ and $0.713$ over all $5{,}040$ orderings. On CUB-200, where 200 classes give far more power, it is $+0.110$ and $+0.081$, with permutation $p$ of $0.123$ and $0.246$ from 20,000 resamples. Neither dataset shows a size effect. RAO is the clearest counterexample: at 19 training images it is the smallest class in the study and the least forgotten of all, with 87.5\% of its samples never forgotten. One qualification belongs with the OCTDL half of this. Its runs equalise class exposure by inverse-frequency weighting (Section~\ref{sec:setup_train}), which is the standard remedy for imbalance but also removes the most obvious route by which class size could act on retention. That null therefore holds for training that already compensates for imbalance, not for unweighted training. CUB-200 carries no such weighting and gives the comparison far more power at 200 classes, and it shows no size effect either. With seven classes, one high-$\lambda$ small class is not evidence of a trend.

The CUB-200 rankings are more suggestive. Its most-forgotten classes under ResNet-18 are California Gull (0.932), Warbling Vireo (0.786), Forsters Tern (0.773) and Blue Grosbeak (0.765), each with close visual analogues elsewhere in the label set. At the other end sit Geococcyx, Blue Jay, Pied Kingfisher and both mergansers, all at $\lambda = 0.000$ and all visually distinctive. The pattern is consistent with confusability driving retention, but we did not test it: establishing it would require a similarity measure defined independently of the model, such as the CUB attribute annotations or a held-out embedding, and we make no quantitative claim here.

\begin{table}[!h]
\centering
\caption{Per-class forgetting on OCTDL under patient-grouped splitting. Each entry is the mean over the three split-varying seeds of that seed's per-class mean, and training sizes are the mean per-seed size, rounded. Sizes vary by seed because patients contribute different numbers of images.}\label{tab:octdl_class}
\resizebox{\columnwidth}{!}{
\begin{tabular}{|l|c|c|c|c|c|}
\hline
& & \multicolumn{2}{c|}{\textbf{ResNet-18}} & \multicolumn{2}{c|}{\textbf{DeiT-Small}} \\
\cline{3-6}
\textbf{Class} & \textbf{Train size} & \textbf{mean $\lambda$} & \textbf{\% never forg.} & \textbf{mean $\lambda$} & \textbf{\% never forg.} \\
\hline
RVO & 72  & 0.876 & 34.2 & 0.845 & 42.2 \\
\hline
ERM & 110 & 0.222 & 46.4 & 0.282 & 53.7 \\
\hline
DME & 102 & 0.136 & 46.2 & 0.108 & 66.9 \\
\hline
VID & 54  & 0.126 & 65.4 & 0.315 & 69.7 \\
\hline
NO  & 232 & 0.086 & 63.6 & 0.065 & 61.9 \\
\hline
AMD & 858 & 0.084 & 58.0 & 0.056 & 87.1 \\
\hline
RAO & 19  & 0.001 & 87.5 & 0.006 & 91.0 \\
\hline
\end{tabular}
}
\end{table}

\subsection{Early loss as a forgetting predictor}\label{sec:res_predictor}

\begin{figure}[!h]
  \centering
  \includegraphics[width=\linewidth]{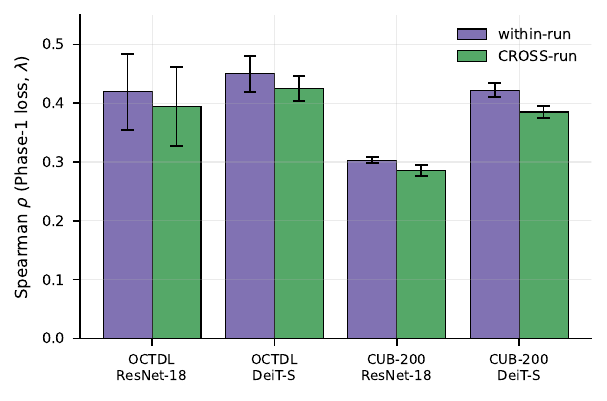}
  \caption{Phase~1 loss against fitted $\lambda$, within-run and across runs. The cross-run version breaks the circularity of measuring both on the same trajectory.}
  \label{fig:predictor}
\end{figure}

Cross-entropy loss at the end of Phase~1, five epochs in and with the backbone still frozen, correlates with the decay constant fitted over the whole run at $\rho = 0.419$ and $0.450$ on OCTDL and $0.303$ and $0.422$ on CUB-200 (Figure~\ref{fig:predictor}). Samples that are hard once the head has warmed up tend to stay hard.

The within-run version of this measurement is partly circular, since the same run produces both the loss and the $\lambda$. Taking the Phase~1 loss from one run and the decay constant from a different seed, joined by filepath, removes the shared-trajectory overlap and barely moves the numbers: $0.395$, $0.425$, $0.286$ and $0.385$ over six ordered pairs drawn from three runs. This does not decouple the two quantities entirely, since $\lambda$ is fitted to the trace after $e^{*}_i$ and Phase~1 loss also predicts $e^{*}_i$, at $\rho = 0.58$ to $0.74$. Partialling $e^{*}_i$ out of both puts the correlation at $+0.521$, $+0.432$, $+0.327$ and $+0.420$ for the four configurations, higher than the raw value in two and marginally lower in the other two. The association is therefore not routed through when a sample is first learned: removing that channel leaves it intact, and in two configurations slightly stronger. Losing only about 0.03 of correlation between the within-run and cross-run versions, a separate comparison from the partialling just described, means the signal is a property of the sample rather than of one run's trajectory.

A quantity available after five frozen-backbone epochs predicts one that otherwise costs a full fine-tuning run to measure, and it does so across runs, which makes it cheap enough to use as a triage step before committing to fine-tuning. It is also the result that makes the sampler's failure informative rather than merely disappointing, since it establishes that the ranking the sampler consumes carries genuine signal.

\subsection{Sampling strategy comparison}\label{sec:res_sampler}

\begin{figure*}[!h]
  \centering
  \includegraphics[width=\textwidth]{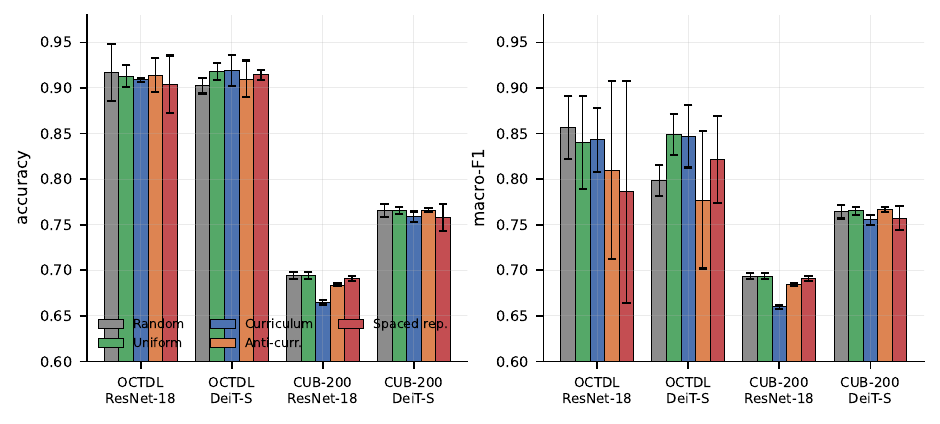}
  \caption{Accuracy and macro F1 for all five arms with bootstrap intervals. The uniform baseline is marked; no arm separates from it after correction.}
  \label{fig:samplingmain}
\end{figure*}

No sampling arm beats uniform. Across five arms, two datasets, two backbones and two metrics, 48 comparisons against the uniform baseline yield a smallest raw $p$ of 0.007 and a smallest Benjamini-Hochberg $q$ of 0.226 (Table~\ref{tab:app_battery}). Nothing survives correction (Table~\ref{tab:main}, Figure~\ref{fig:samplingmain}). The largest effects run the wrong way: curriculum sampling loses 3.4 points of macro F1 on CUB-200 with ResNet-18, and anti-curriculum loses 1.0 points of accuracy on the same configuration.

At the sample level, McNemar's test does find differences. This is a separate family, also of size 48, testing individual test images rather than seed means. Five of its 48 comparisons survive correction (Table~\ref{tab:app_mcnemar}), all on CUB-200, and all in the direction of uniform: curriculum loses in all three seeds, anti-curriculum in one, spaced repetition in one. The arms are changing which images are classified correctly. They are not changing how many.

\begin{figure*}[!h]
  \centering
  \includegraphics[width=\textwidth]{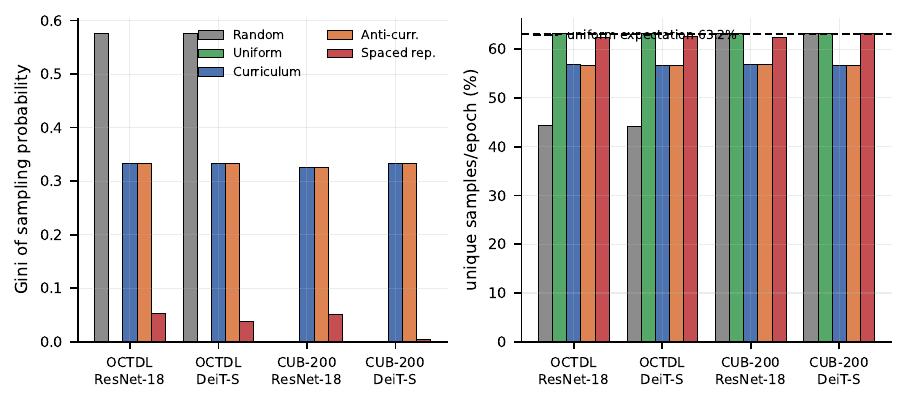}
  \caption{Sampling diagnostics per arm: Gini coefficient, unique-sample fraction and max/min probability ratio. These measure whether an arm prioritised, independently of whether it helped.}
  \label{fig:diagnostics}
\end{figure*}

The reason the spaced-repetition arm does nothing is visible in its own diagnostics rather than in its accuracy. At $\tau = 1.0$ the measured ratio between the largest and smallest sampling probability is 2.718, matching the bound $\exp(1/\tau)$ from \eqref{eq:bound} to three decimals. On OCTDL with ResNet-18 the arm draws 62.3\% unique samples per epoch against uniform sampling's 63.2\%, and its Gini coefficient is 0.053, lower than curriculum's 0.333 and lower than the class-balanced random arm's 0.575. The DeiT-Small figures are 62.6\% and 0.037 (Figure~\ref{fig:diagnostics}). By any of these measures the sampler was not prioritising anything, so the null result at $\tau = 1$ carries no information about spaced repetition.

\begin{figure*}[!h]
  \centering
  \includegraphics[width=\textwidth]{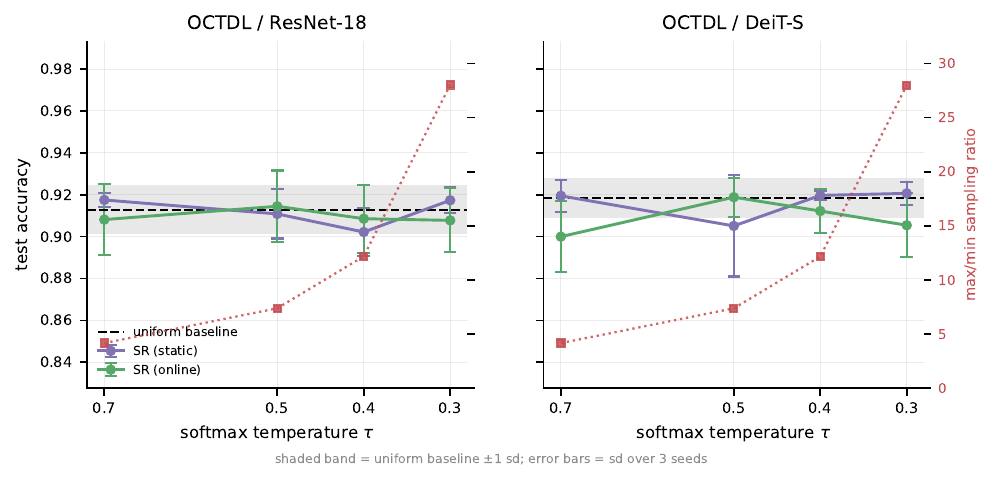}
  \caption{Accuracy against temperature on OCTDL, with the uniform baseline as a shaded band. Prioritisation strength rises 28-fold from left to right while accuracy stays inside the band.}
  \label{fig:tausweep}
\end{figure*}

Sweeping $\tau$ down to 0.3 fixes the sampler and leaves the conclusion unchanged. The probability ratio rises to 4.2$\times$, 7.4$\times$, 12.2$\times$ and 28.0$\times$, tracking $\exp(1/\tau)$ across the whole grid, and across the same four temperatures, unique-sample coverage falls from 61.0\% to 45.0\% for ResNet-18 and from 61.9\% to 51.3\% for DeiT-Small (Table~\ref{tab:tau}, Figure~\ref{fig:tausweep}). By the time $\tau = 0.3$ the sampler is genuinely concentrating on the samples it judges most fragile. Accuracy does not respond: ResNet-18 moves between 0.902 and 0.917 across the grid with no monotone trend, DeiT-Small between 0.905 and 0.921, and the uniform baselines sit at 0.913 and 0.918, inside both ranges. Adding the online variant, which refits $\lambda$ during training instead of using constants from a prior run, changes nothing. All 16 $\tau$-sweep comparisons fail correction.

That combination is what makes the negative result worth reporting. The sampler prioritises, by a factor of 28. Its ranking is reproducible across seeds at $\rho \approx 0.37$ to $0.59$ (Section~\ref{sec:res_stability}) and predictable from Phase~1 loss across runs at $\rho \approx 0.29$ to $0.43$ (Section~\ref{sec:res_predictor}). Exposure is matched across arms. Under those conditions, reweighting training exposure by estimated fragility still does not improve generalisation. An unstable per-sample signal would be the natural explanation, and it is not available here: the signal is stable and the null persists.

This is also where our result and Kline's~\cite{kline2025human} stop being comparable. His spaced reviews restore a class whose examples had been withdrawn from training, so there is something to reinstate; ours redistribute exposure among samples that were never withdrawn. A schedule that reintroduces absent data and a schedule that reweights present data are different interventions, and only the first has an obvious deficit to repair. The transition rates above suggest why the second does not: a model that relearns a present sample with probability $0.47$ to $0.70$ has little to recover.

We think the more likely explanation is that per-sample retention and generalisation are less tightly coupled than the memory analogy suggests. A sample that oscillates between correct and incorrect on the training set is not necessarily a sample the model needs to see more often, and with $p_\text{relearn}$ between 0.47 and 0.70 the model recovers such samples on its own within an epoch or two. Spaced repetition helps human learners because forgetting is costly and recovery is slow. Neither holds here.

\begin{table*}[!h]
\centering
\caption{Classification performance across the five sampling arms at $\tau=1.0$ (mean $\pm$ sd, $\mathrm{ddof}=1$, 3 seeds). On CUB with ResNet-18 the Random and Uniform arms are identical by construction, since that dataset is balanced and the two draw the same way; with DeiT-Small they differ slightly despite being uniform in expectation. No arm differs from Uniform after Benjamini-Hochberg correction.}\label{tab:main}
\resizebox{\textwidth}{!}{
\begin{tabular}{|l|l|c|c|c|c|c|c|c|c|c|c|}
\hline
& & \multicolumn{5}{c|}{\textbf{Accuracy}} & \multicolumn{5}{c|}{\textbf{Macro F1}} \\
\cline{3-12}
\textbf{Dataset} & \textbf{Backbone} & \textbf{Random} & \textbf{Uniform} & \textbf{Curriculum} & \textbf{Anti-curr.} & \textbf{Spaced rep.} & \textbf{Random} & \textbf{Uniform} & \textbf{Curriculum} & \textbf{Anti-curr.} & \textbf{Spaced rep.} \\
\hline
OCTDL & ResNet-18 & \textbf{0.917$\pm$0.031} & 0.913$\pm$0.012 & 0.909$\pm$0.002 & 0.914$\pm$0.018 & 0.904$\pm$0.031 & \textbf{0.856$\pm$0.035} & 0.840$\pm$0.051 & 0.843$\pm$0.035 & 0.810$\pm$0.098 & 0.786$\pm$0.122 \\
\hline
OCTDL & DeiT-S & 0.902$\pm$0.008 & 0.918$\pm$0.009 & \textbf{0.919$\pm$0.017} & 0.910$\pm$0.020 & 0.914$\pm$0.006 & 0.798$\pm$0.017 & \textbf{0.849$\pm$0.023} & 0.847$\pm$0.034 & 0.777$\pm$0.075 & 0.822$\pm$0.048 \\
\hline
CUB-200 & ResNet-18 & \textbf{0.694$\pm$0.004} & \textbf{0.694$\pm$0.004} & 0.665$\pm$0.002 & 0.684$\pm$0.002 & 0.691$\pm$0.003 & \textbf{0.694$\pm$0.003} & \textbf{0.694$\pm$0.003} & 0.660$\pm$0.002 & 0.685$\pm$0.002 & 0.691$\pm$0.003 \\
\hline
CUB-200 & DeiT-S & 0.766$\pm$0.007 & 0.766$\pm$0.004 & 0.759$\pm$0.006 & \textbf{0.766$\pm$0.002} & 0.758$\pm$0.015 & 0.764$\pm$0.008 & 0.765$\pm$0.004 & 0.755$\pm$0.005 & \textbf{0.766$\pm$0.003} & 0.757$\pm$0.013 \\
\hline
\end{tabular}
}
\end{table*}

\begin{table*}[!h]
\centering
\caption{Temperature sweep on OCTDL. The measured probability ratio, shown for ResNet-18, tracks the bound $\exp(1/\tau)$ from \eqref{eq:bound} across the whole grid while accuracy stays inside the uniform baseline's range. DeiT-Small's realised ratios follow the same bound, differing from ResNet-18's by at most $0.71$ in those ratio units, at the low-temperature end of the online sweep. Mean over 3 seeds.}\label{tab:tau}
\resizebox{\textwidth}{!}{
\begin{tabular}{|l|c|c|c|c|c|}
\hline
\textbf{Arm} & \textbf{$\tau$} & \textbf{ratio (ResNet-18)} & \textbf{$\exp(1/\tau)$} & \textbf{acc (ResNet-18)} & \textbf{acc (DeiT-S)} \\
\hline
Uniform & -- & 1.00$\times$ & -- & 0.913 & 0.918 \\
\hline
Spaced rep. & 1.0 & 2.72$\times$ & 2.72$\times$ & 0.904 & 0.914 \\
\hline
Spaced rep. & 0.7 & 4.17$\times$ & 4.17$\times$ & 0.917 & 0.919 \\
\hline
Spaced rep. & 0.5 & 7.39$\times$ & 7.39$\times$ & 0.911 & 0.905 \\
\hline
Spaced rep. & 0.4 & 12.18$\times$ & 12.18$\times$ & 0.902 & 0.920 \\
\hline
Spaced rep. & 0.3 & 28.02$\times$ & 28.03$\times$ & 0.917 & 0.921 \\
\hline
Spaced rep. (online) & 0.7 & 4.17$\times$ & 4.17$\times$ & 0.908 & 0.900 \\
\hline
Spaced rep. (online) & 0.5 & 7.38$\times$ & 7.39$\times$ & 0.914 & 0.919 \\
\hline
Spaced rep. (online) & 0.4 & 12.17$\times$ & 12.18$\times$ & 0.909 & 0.912 \\
\hline
Spaced rep. (online) & 0.3 & 28.01$\times$ & 28.03$\times$ & 0.908 & 0.905 \\
\hline
\end{tabular}
}
\end{table*}

\subsection{Discussion}\label{sec:res_discussion}

Three of the results above correct our own earlier analysis, and they fail in the same way. Positional alignment across runs with different data ordering, $R^2$ scored on constant traces, and raw Jaccard overlap read without a chance baseline each produce plausible numbers with no error and no warning. The defect in every case is a missing null: what the statistic reads when the effect is absent, which costs a few lines of code to compute and would have caught all three.

What survives is a narrower picture, and a better supported one. Per-sample retention during fine-tuning is reproducible enough to be treated as a property of the sample. It differs measurably between a CNN and a ViT trained on identical data. A cheap early-training signal predicts it, and does so across runs. None of that makes it a useful control variable for training, at least not through sampling frequency: with prioritisation swept over a 28-fold range and exposure matched, no arm beat uniform on either dataset.

The exponential decay model should be dropped, and so should its monotone relatives. It fits worse than a per-sample mean on the traces that actually contain forgetting events, a free-asymptote variant and a power law do no better on the same traces (Table~\ref{tab:altmodels}), and the two-state Markov estimates suggest why: forgetting during fine-tuning is a low-rate, high-recovery process, not a decay. The one thing $\lambda$ does reliably encode, whether a sample is ever forgotten at all, is captured more directly by counting forgetting events, which requires no curve fitting and has no undefined cases. Future work on retention dynamics would be better served by that statistic, or by a two-state model fitted to the transitions.

Scale and training regime bound all of this, and Section~\ref{sec:limitations} sets out both.

\section{Conclusion}\label{sec:conclusion}

Three claims survive this study, and each is narrower than the version we reported previously.

Per-sample retention during fine-tuning is a property of the sample rather than of the run. Two independently seeded runs of one architecture on one split agree on which samples are fragile at $\rho = 0.37$ to $0.59$, and the first-learned epoch is more reproducible still, reaching $0.72$. Curriculum learning, data pruning and dataset cartography all compute a score from a single trajectory and then act on it, and in this regime that score does transfer. Our earlier report that it does not was an artifact of matching samples by position in the training array while the split changed underneath.

Architecture matters, by a measurable but modest amount. With the split held fixed, within-architecture agreement exceeds between-architecture agreement by $0.145$ on OCTDL and $0.173$ on CUB-200, and the top-decile overlap between the CNN and the ViT runs $5.6\times$ and $2.7\times$ above its chance value. The two backbones neither agree nor disagree completely about which samples are hard, which cautions against treating either one's difficulty ranking as the ranking.

Five epochs with the backbone frozen produce a score that predicts the decay constant of a separate training run at $\rho = 0.29$ to $0.43$, at a small fraction of the cost of measuring the decay constant directly.

Set against those, two negatives. The exponential decay model does not describe the traces it is fitted to: restricted to traces containing at least one forgetting event, mean $R^2$ is negative in all four configurations, meaning the fitted curve predicts a trace worse than that sample's own mean. Sorted by shape, the reason is direct: monotone decay, the one shape the model represents, is $0.1\%$ to $0.8\%$ of samples in every configuration. Fitting a free asymptote or a power law instead does not help, which places the failure in monotonicity rather than in the particular curve. Forgetting here is a low-rate, high-recovery process, with a per-epoch forgetting probability of $0.020$ to $0.051$ against a relearning probability of $0.47$ to $0.70$, and no monotone curve can represent that. Scheduling on the decay constants does not work either: with prioritisation swept over a 28-fold range, exposure matched across arms, and an online variant that re-estimates the constants during training, none of 48 comparisons against uniform sampling survives correction.

A third negative is more provisional. Patient-grouped splitting is the standard remedy for the leakage in Section~\ref{sec:setup_split}, and on OCTDL it changes the headline metrics by less than their run-to-run spread, even though three quarters or more of the test set shares a patient with training. The leak is real and the remedy is still correct; what it does not appear to buy, here, is the accuracy correction it is usually assumed to.

For practice the two negatives matter more than they might appear to. A pipeline that computes per-sample difficulty once and reweights training exposure by it has a stable and inexpensive score to work from, and still gains nothing. What needs rethinking is therefore not how we estimate which samples are fragile, but what we do with the estimate.

\subsection{Limitations and future work}\label{sec:limitations}

The sampling comparison is by far the least powered part of the study, and it is worth being exact about how little it can rule out. At three seeds per arm a paired test reaches 80\% power only at a raw standardised effect of $d_z \approx 3.3$, equivalently a bias-corrected Hedges' $g$ of about $1.9$, the unit reported in Table~\ref{tab:app_battery}. Against the mean standard deviation of the paired accuracy difference across our comparisons, $0.012$, that is a gap of roughly four accuracy points, and correcting 48 comparisons for multiplicity raises the threshold further. Nothing smaller could have been detected whatever its size, so the null is evidence against a large effect and says little about a small one. The temperature sweep and the sample-level McNemar tests point the same way, which is reassurance rather than power. The retention analyses rest on ten seeds and are correspondingly firmer.

Scale and scope bound the rest. Both datasets are small, at 1,428 and 5,094 training images, and retention structure in a model trained on orders of magnitude more data may differ. We compare one CNN against one ViT, so "architecture-specific" means specific to that pair rather than to CNNs and ViTs in general. The horizon difference between the two backbones is controlled by the truncated measurement in Section~\ref{sec:res_arch}, but the smaller batch size used for DeiT-Small on CUB-200 is not, and separating it would need retraining. A second asymmetry is not controlled either. Retention is recorded in inference mode, where ResNet-18's batch normalisation layers use running statistics that lag the current weights, while DeiT-Small's layer normalisation is computed per sample and carries no such lag. Recorded correctness for ResNet-18 may therefore carry epoch-to-epoch jitter that reflects stale normalisation rather than anything the network has learned, and that jitter would push in the direction we report, since ResNet-18 is the backbone with more forgetting events. Telling the two apart would require re-running ResNet-18 with its batch statistics recomputed before each evaluation. Every result concerns fine-tuning from ImageNet-pretrained weights, where the backbone already encodes most of what the task needs, and training from scratch is a different regime we did not test. The ResNet-18 runs on CUB-200 reach the 50-epoch budget on every seed and are genuinely truncated rather than merely capped: across the thirteen such runs the best validation loss falls between epochs 43 and 49, and in eight of them within the final five epochs, so the ceiling bound while the model was still improving. The fixed-horizon reanalyses in Sections~\ref{sec:res_char} and~\ref{sec:res_arch} show that neither the constant-trace gap nor the architecture gap is an artifact of it.

Our leakage comparison is the most provisional result here, for the reasons set out alongside it in Section~\ref{sec:res_char}: it establishes that the correction is small on this dataset, not that leakage is generally harmless.

Two claims we make qualitatively deserve a quantitative test. The CUB-200 species that are forgotten most have close visual analogues in the label set and those forgotten least are distinctive, but we did not measure similarity independently of the model, which the attribute annotations or a held-out embedding would allow. And the observation that two architectures forget different samples suggests they would make complementary ensemble members, which the saved test predictions would let someone check directly.

Three directions follow from the negatives. The decay constant should be replaced: what it reliably encodes is whether a sample is ever forgotten, which a forgetting-event count captures with no curve fitting and no undefined cases, and the transition structure is better served by a two-state model fitted to the trace directly. Sampling frequency is only one way to act on a difficulty score, and loss reweighting, targeted augmentation or selective pruning may use the same stable signal to better effect. Finally, mean decay varies substantially between classes even though it does not track class size, from $0.876$ for RVO down to $0.001$ for RAO, so scheduling at the class level on measured forgetting rather than on class frequency remains open.

\section*{Declaration of Competing Interest}
The authors declare that they have no known competing financial interests or personal relationships that could have appeared to influence the work reported in this paper.

\section*{Acknowledgments}
This work was supported by the open access funding provided by Vellore Institute of Technology, Vellore.

\section*{Data Availability}
The datasets analyzed in this study are publicly available.
OCTDL is available at \href{https://www.kaggle.com/datasets/shakilrana/octdl-retinal-oct-images-dataset}{Kaggle}.
CUB-200-2011 is available at \href{https://www.kaggle.com/datasets/wenewone/cub2002011}{Kaggle}.
The code used in this study is available from the corresponding author upon reasonable request.

\appendix
\section{Per-class test metrics on OCTDL}\label{app:perclass}

Table~\ref{tab:app_perclass} gives per-class precision, recall, specificity and F1 on the OCTDL test set, averaged over the three split-varying seeds under patient-grouped splitting. Support is the mean number of test images per class. Note that RAO contributes about one test image per split, so its per-class figures carry no useful precision and are shown only for completeness. CUB-200 has 200 classes and is not tabulated per class; its macro-averaged values appear in Table~\ref{tab:main}.

\begin{table*}[!h]
\centering
\caption{Per-class test metrics on OCTDL, mean over three split-varying seeds. RAO has a mean support of one image and its row is not interpretable.}\label{tab:app_perclass}

\begin{tabular}{|l|c|c|c|c|c|c|c|c|c|}
\hline
& & \multicolumn{4}{c|}{\textbf{ResNet-18}} & \multicolumn{4}{c|}{\textbf{DeiT-Small}} \\
\cline{3-10}
\textbf{Class} & \textbf{Support} & \textbf{Prec.} & \textbf{Rec.} & \textbf{Spec.} & \textbf{F1} & \textbf{Prec.} & \textbf{Rec.} & \textbf{Spec.} & \textbf{F1} \\
\hline
AMD & 186 & 0.958 & 0.969 & 0.935 & 0.963 & 0.964 & 0.956 & 0.940 & 0.960 \\
\hline
DME & 22 & 0.763 & 0.746 & 0.980 & 0.748 & 0.701 & 0.819 & 0.970 & 0.747 \\
\hline
ERM & 23 & 0.835 & 0.810 & 0.987 & 0.822 & 0.848 & 0.819 & 0.987 & 0.831 \\
\hline
NO & 51 & 0.904 & 0.927 & 0.981 & 0.915 & 0.857 & 0.886 & 0.971 & 0.871 \\
\hline
RAO & 1 & 1.000 & 1.000 & 1.000 & 1.000 & 0.833 & 1.000 & 0.999 & 0.889 \\
\hline
RVO & 17 & 0.768 & 0.640 & 0.991 & 0.695 & 0.869 & 0.638 & 0.994 & 0.728 \\
\hline
VID & 10 & 0.849 & 0.854 & 0.994 & 0.849 & 0.933 & 0.738 & 0.997 & 0.809 \\
\hline
\end{tabular}

\end{table*}

\section{Full statistical battery}\label{app:stats}

Table~\ref{tab:app_battery} lists every seed-level comparison of a sampling arm against the uniform baseline: the paired difference, the paired $t$-test $p$-value, the Wilcoxon $p$-value, Hedges' $g$, and the Benjamini-Hochberg corrected $q$. All 48 rows are shown, ordered by raw $p$; none survives correction at $q=0.05$. The Wilcoxon column is bounded below by $0.25$ at three seeds, as noted in Section~\ref{sec:setup_stats}. Table~\ref{tab:app_mcnemar} gives the corresponding sample-level McNemar tests.

\begin{table*}[!h]
\centering
\caption{All 48 seed-level comparisons against uniform sampling, ordered by raw $p$ down the left block and continuing down the right, with undefined rows last. None survives Benjamini-Hochberg correction. The two undefined rows are CUB-200 with ResNet-18 under the Random arm, where Random and Uniform coincide exactly on every seed because that dataset is balanced, so the paired difference is identically zero and no test statistic is defined.}\label{tab:app_battery}
\resizebox{\textwidth}{!}{
\begin{tabular}{|l|l|l|l|c|c|c|c|c||l|l|l|l|c|c|c|c|c|}
\hline
\textbf{Data} & \textbf{Net} & \textbf{Arm} & \textbf{Met.} & \textbf{$\Delta$} & \textbf{$p$} & \textbf{Wilc.} & \textbf{$g$} & \textbf{$q$} & \textbf{Data} & \textbf{Net} & \textbf{Arm} & \textbf{Met.} & \textbf{$\Delta$} & \textbf{$p$} & \textbf{Wilc.} & \textbf{$g$} & \textbf{$q$} \\
\hline
CUB & R18 & curr & F1 & $-0.0336$ & 0.007 & 0.25 & $-3.81$ & 0.226 & OCTDL & R18 & SR $\tau$0.7 & acc & $+0.0046$ & 0.576 & -- & $+0.22$ & 1.000 \\
\hline
CUB & R18 & curr & acc & $-0.0295$ & 0.012 & 0.25 & $-2.95$ & 0.226 & OCTDL & R18 & SR & acc & $-0.0087$ & 0.628 & 0.75 & $-0.19$ & 1.000 \\
\hline
CUB & R18 & anti-curr & acc & $-0.0104$ & 0.021 & 0.25 & $-2.26$ & 0.226 & CUB & DeiT & anti-curr & F1 & $+0.0011$ & 0.632 & 1.00 & $+0.18$ & 1.000 \\
\hline
CUB & R18 & anti-curr & F1 & $-0.0090$ & 0.022 & 0.25 & $-2.18$ & 0.226 & OCTDL & R18 & SR onl. $\tau$0.3 & acc & $-0.0051$ & 0.632 & -- & $-0.18$ & 1.000 \\
\hline
CUB & DeiT & curr & F1 & $-0.0099$ & 0.033 & 0.25 & $-1.76$ & 0.226 & OCTDL & R18 & random & F1 & $+0.0164$ & 0.633 & 0.75 & $+0.18$ & 1.000 \\
\hline
OCTDL & DeiT & random & F1 & $-0.0509$ & 0.035 & 0.25 & $-1.71$ & 0.226 & OCTDL & DeiT & SR & acc & $-0.0040$ & 0.643 & 0.75 & $-0.18$ & 1.000 \\
\hline
OCTDL & R18 & SR $\tau$0.4 & acc & $-0.0107$ & 0.036 & -- & $-1.70$ & 0.226 & OCTDL & R18 & SR $\tau$0.5 & acc & $-0.0021$ & 0.681 & -- & $-0.16$ & 1.000 \\
\hline
CUB & DeiT & curr & acc & $-0.0066$ & 0.039 & 0.25 & $-1.61$ & 0.226 & OCTDL & R18 & anti-curr & F1 & $-0.0301$ & 0.707 & 0.75 & $-0.14$ & 1.000 \\
\hline
OCTDL & DeiT & random & acc & $-0.0162$ & 0.042 & 0.25 & $-1.55$ & 0.226 & OCTDL & R18 & SR onl. $\tau$0.7 & acc & $-0.0047$ & 0.776 & -- & $-0.11$ & 1.000 \\
\hline
OCTDL & DeiT & SR onl. $\tau$0.7 & acc & $-0.0184$ & 0.066 & -- & $-1.22$ & 0.319 & OCTDL & DeiT & SR $\tau$0.3 & acc & $+0.0022$ & 0.801 & -- & $+0.09$ & 1.000 \\
\hline
OCTDL & DeiT & anti-curr & F1 & $-0.0720$ & 0.147 & 0.25 & $-0.76$ & 0.642 & OCTDL & R18 & random & acc & $+0.0041$ & 0.802 & 0.75 & $+0.09$ & 1.000 \\
\hline
CUB & R18 & SR & acc & $-0.0034$ & 0.199 & 0.50 & $-0.62$ & 0.796 & CUB & DeiT & anti-curr & acc & $+0.0007$ & 0.806 & 1.00 & $+0.09$ & 1.000 \\
\hline
OCTDL & DeiT & SR & F1 & $-0.0276$ & 0.250 & 0.25 & $-0.53$ & 0.922 & OCTDL & DeiT & curr & F1 & $-0.0022$ & 0.808 & 1.00 & $-0.09$ & 1.000 \\
\hline
CUB & R18 & SR & F1 & $-0.0025$ & 0.295 & 0.50 & $-0.46$ & 0.925 & CUB & DeiT & random & F1 & $-0.0008$ & 0.838 & 1.00 & $-0.08$ & 1.000 \\
\hline
OCTDL & DeiT & SR $\tau$0.5 & acc & $-0.0133$ & 0.311 & -- & $-0.44$ & 0.925 & OCTDL & DeiT & SR $\tau$0.4 & acc & $+0.0013$ & 0.854 & -- & $+0.07$ & 1.000 \\
\hline
OCTDL & DeiT & SR onl. $\tau$0.4 & acc & $-0.0062$ & 0.314 & -- & $-0.44$ & 0.925 & OCTDL & R18 & curr & F1 & $+0.0031$ & 0.870 & 1.00 & $+0.06$ & 1.000 \\
\hline
OCTDL & DeiT & anti-curr & acc & $-0.0086$ & 0.328 & 0.50 & $-0.42$ & 0.925 & OCTDL & DeiT & curr & acc & $+0.0008$ & 0.904 & 1.00 & $+0.04$ & 1.000 \\
\hline
OCTDL & R18 & SR $\tau$0.3 & acc & $+0.0045$ & 0.369 & -- & $+0.38$ & 0.982 & OCTDL & DeiT & SR $\tau$0.7 & acc & $+0.0010$ & 0.905 & -- & $+0.04$ & 1.000 \\
\hline
OCTDL & DeiT & SR onl. $\tau$0.3 & acc & $-0.0130$ & 0.389 & -- & $-0.36$ & 0.982 & OCTDL & R18 & anti-curr & acc & $+0.0013$ & 0.915 & 1.00 & $+0.04$ & 1.000 \\
\hline
CUB & DeiT & SR & F1 & $-0.0082$ & 0.442 & 0.50 & $-0.31$ & 1.000 & OCTDL & R18 & SR onl. $\tau$0.5 & acc & $+0.0016$ & 0.918 & -- & $+0.04$ & 1.000 \\
\hline
OCTDL & R18 & SR onl. $\tau$0.4 & acc & $-0.0043$ & 0.463 & -- & $-0.30$ & 1.000 & OCTDL & DeiT & SR onl. $\tau$0.5 & acc & $+0.0004$ & 0.970 & -- & $+0.01$ & 1.000 \\
\hline
CUB & DeiT & SR & acc & $-0.0073$ & 0.524 & 0.50 & $-0.25$ & 1.000 & CUB & DeiT & random & acc & $+0.0000$ & 1.000 & 1.00 & $-0.00$ & 1.000 \\
\hline
OCTDL & R18 & curr & acc & $-0.0040$ & 0.557 & 1.00 & $-0.23$ & 1.000 & CUB & R18 & random & acc & $+0.0000$ & -- & 1.00 & -- & 1.000 \\
\hline
OCTDL & R18 & SR & F1 & $-0.0540$ & 0.564 & 0.75 & $-0.23$ & 1.000 & CUB & R18 & random & F1 & $+0.0000$ & -- & 1.00 & -- & 1.000 \\
\hline
\end{tabular}
}
\end{table*}

\begin{table*}[!h]
\centering
\caption{All 48 sample-level McNemar comparisons against uniform sampling, ordered by raw $p$ down the left block and continuing down the right. ``Unif.\ only'' counts test images the uniform arm classifies correctly and the compared arm does not; ``arm only'' is the reverse. Rows marked $\ast$ survive Benjamini-Hochberg correction, and every one of them favours uniform.}\label{tab:app_mcnemar}
\resizebox{\textwidth}{!}{
\begin{tabular}{|l|l|l|c|c|c|c|c|c||l|l|l|c|c|c|c|c|c|}
\hline
\textbf{Data} & \textbf{Net} & \textbf{Arm} & \textbf{Seed} & \textbf{U. only} & \textbf{A. only} & \textbf{$\chi^2$} & \textbf{$p$} & \textbf{$q$} & \textbf{Data} & \textbf{Net} & \textbf{Arm} & \textbf{Seed} & \textbf{U. only} & \textbf{A. only} & \textbf{$\chi^2$} & \textbf{$p$} & \textbf{$q$} \\
\hline
CUB & R18 & curr & 2026 & 398 & 190 & 72.87 & 0.000 & 0.000$\ast$ & OCTDL & R18 & SR & 2026 & 7 & 12 & 0.84 & 0.359 & 0.689 \\
\hline
CUB & R18 & curr & 42 & 344 & 183 & 48.58 & 0.000 & 0.000$\ast$ & CUB & DeiT & curr & 2026 & 442 & 415 & 0.79 & 0.374 & 0.691 \\
\hline
CUB & R18 & curr & 99 & 367 & 223 & 34.66 & 0.000 & 0.000$\ast$ & OCTDL & DeiT & curr & 42 & 5 & 9 & 0.64 & 0.423 & 0.751 \\
\hline
CUB & DeiT & SR & 42 & 483 & 357 & 18.60 & 0.000 & 0.000$\ast$ & OCTDL & DeiT & random & 42 & 10 & 6 & 0.56 & 0.453 & 0.777 \\
\hline
CUB & R18 & anti-curr & 2026 & 292 & 215 & 11.39 & 0.001 & 0.007$\ast$ & OCTDL & DeiT & random & 99 & 14 & 10 & 0.38 & 0.540 & 0.874 \\
\hline
CUB & R18 & anti-curr & 99 & 292 & 235 & 5.95 & 0.015 & 0.105 & OCTDL & R18 & anti-curr & 42 & 7 & 4 & 0.36 & 0.546 & 0.874 \\
\hline
OCTDL & R18 & SR & 99 & 14 & 3 & 5.88 & 0.015 & 0.105 & OCTDL & R18 & random & 42 & 5 & 8 & 0.31 & 0.579 & 0.892 \\
\hline
CUB & DeiT & SR & 2026 & 340 & 402 & 5.01 & 0.025 & 0.151 & OCTDL & R18 & anti-curr & 99 & 9 & 6 & 0.27 & 0.606 & 0.892 \\
\hline
CUB & DeiT & SR & 99 & 474 & 411 & 4.34 & 0.037 & 0.197 & CUB & DeiT & anti-curr & 42 & 391 & 376 & 0.26 & 0.613 & 0.892 \\
\hline
CUB & R18 & anti-curr & 42 & 277 & 230 & 4.17 & 0.041 & 0.197 & OCTDL & DeiT & anti-curr & 42 & 11 & 8 & 0.21 & 0.646 & 0.911 \\
\hline
CUB & DeiT & random & 2026 & 261 & 219 & 3.50 & 0.061 & 0.267 & OCTDL & DeiT & SR & 99 & 10 & 13 & 0.17 & 0.677 & 0.911 \\
\hline
OCTDL & R18 & random & 2026 & 4 & 12 & 3.06 & 0.080 & 0.306 & OCTDL & R18 & SR & 42 & 4 & 2 & 0.17 & 0.683 & 0.911 \\
\hline
CUB & DeiT & curr & 99 & 476 & 423 & 3.01 & 0.083 & 0.306 & OCTDL & R18 & curr & 2026 & 8 & 8 & 0.06 & 0.803 & 1.000 \\
\hline
CUB & R18 & SR & 2026 & 190 & 158 & 2.76 & 0.097 & 0.331 & OCTDL & DeiT & SR & 2026 & 9 & 7 & 0.06 & 0.803 & 1.000 \\
\hline
OCTDL & R18 & anti-curr & 2026 & 4 & 11 & 2.40 & 0.121 & 0.388 & CUB & DeiT & random & 42 & 308 & 314 & 0.04 & 0.841 & 1.000 \\
\hline
CUB & R18 & SR & 42 & 175 & 147 & 2.26 & 0.132 & 0.397 & OCTDL & DeiT & curr & 99 & 18 & 16 & 0.03 & 0.864 & 1.000 \\
\hline
OCTDL & R18 & random & 99 & 12 & 5 & 2.12 & 0.146 & 0.411 & CUB & DeiT & anti-curr & 2026 & 412 & 407 & 0.02 & 0.889 & 1.000 \\
\hline
CUB & DeiT & random & 99 & 353 & 389 & 1.65 & 0.199 & 0.530 & OCTDL & R18 & curr & 99 & 8 & 9 & 0.00 & 1.000 & 1.000 \\
\hline
CUB & DeiT & curr & 42 & 448 & 414 & 1.26 & 0.261 & 0.592 & OCTDL & DeiT & curr & 2026 & 8 & 7 & 0.00 & 1.000 & 1.000 \\
\hline
OCTDL & DeiT & random & 2026 & 18 & 11 & 1.24 & 0.265 & 0.592 & OCTDL & DeiT & anti-curr & 2026 & 7 & 8 & 0.00 & 1.000 & 1.000 \\
\hline
OCTDL & R18 & curr & 42 & 9 & 4 & 1.23 & 0.267 & 0.592 & CUB & R18 & random & 42 & 0 & 0 & 0.00 & 1.000 & 1.000 \\
\hline
CUB & DeiT & anti-curr & 99 & 381 & 413 & 1.21 & 0.271 & 0.592 & CUB & R18 & random & 99 & 0 & 0 & 0.00 & 1.000 & 1.000 \\
\hline
OCTDL & DeiT & SR & 42 & 11 & 6 & 0.94 & 0.332 & 0.689 & CUB & R18 & random & 2026 & 0 & 0 & 0.00 & 1.000 & 1.000 \\
\hline
OCTDL & DeiT & anti-curr & 99 & 17 & 11 & 0.89 & 0.345 & 0.689 & CUB & R18 & SR & 99 & 182 & 183 & 0.00 & 1.000 & 1.000 \\
\hline
\end{tabular}
}
\end{table*}

\printcredits

\bibliographystyle{model1-num-names}

\bibliography{cas-refs}

\end{document}